\documentclass{article}

\usepackage[nonatbib,preprint]{neurips_2022}

\usepackage[utf8]{inputenc} 
\usepackage[T1]{fontenc}    
\usepackage{hyperref}       
\usepackage{url}            
\usepackage{booktabs}       
\usepackage{amsfonts}       
\usepackage{nicefrac}       
\usepackage{microtype}      
\usepackage{xcolor}         

\usepackage{graphicx} 
\graphicspath{ {./Images/} }
\usepackage{caption}
\usepackage{subcaption}
\usepackage{diagbox}
\usepackage{amsmath}
\usepackage{amssymb}
\usepackage{comment}
\usepackage{wrapfig}
\usepackage[square,numbers]{natbib}

\title{Understanding, Detecting, and Separating Out-of-Distribution Samples and Adversarial Samples in Text Classification}

\author{
  Cheng-Han Chiang \\
  National Taiwan University\\
  \texttt{dcml0714@ntu.edu.tw} \\
   \And
   Hung-yi Lee \\
   National Taiwan University \\
   hungyilee@ntu.edu.tw \\
}

\begin{document}

\maketitle

\begin{abstract}
\label{abstract}
In this paper, we study the differences and commonalities between statistically out-of-distribution (OOD) samples and adversarial (Adv) samples, both of which hurting a text classification model's performance.
We conduct analyses to compare the two types of anomalies (OOD and Adv samples) with the in-distribution (ID) ones from three aspects: the input features, the hidden representations in each layer of the model, and the output probability distributions of the classifier.
We find that OOD samples expose their aberration starting from the first layer, while the abnormalities of Adv samples do not emerge until the deeper layers of the model.
We also illustrate that the models' output probabilities for Adv samples tend to be more unconfident.
Based on our observations, we propose a simple method to separate ID, OOD, and Adv samples using the hidden representations and output probabilities of the model.
On multiple combinations of ID, OOD datasets, and Adv attacks, our proposed method shows exceptional results on distinguishing ID, OOD, and Adv samples.
\end{abstract}

\section{Introduction}
\label{sec: intro}
Deep learning-based text classification models have achieved overwhelming success on miscellaneous benchmark datasets~\citep{devlin-etal-2019-bert, liu2019roberta} and real-world applications.
Despite their great success, the performances of those models are shown to degrade when faced with data samples drawn from a distribution that is very different from the training distribution (i.e., in-distribution data), including samples that are statistically out-of-distribution (OOD)~\citep{hendrycks-etal-2020-pretrained}, and maliciously created adversaries (Adv)~\citep{wang-etal-2021-closer,maheshwary2021generating,li-etal-2020-bert-attack}.
In this work, we refer to OOD samples as those samples that are statistically different from the ID dataset.
Adversarial examples refer to those samples created adversarially and intentionally by some wicked individuals, intending to bring down the performance of a trained model.

We argue that for a robust text classifier, it should be able to distinguish between Adv and OOD samples.
Since Adv samples are crafted to harm the model performance while OOD samples are not, they should be treated differently:
if the model detects an input as an Adv example, it can preprocess the input into a non-adversarial example before feeding it into the text classifier~\citep{wang2020defense}; if an input sample is detected as an OOD sample, the model should not make any prediction since the result might not make any sense.
While detecting OOD samples~\citep{xu-etal-2021-unsupervised, hendrycks-etal-2020-pretrained, xu-etal-2020-deep} and detecting Adv samples~\citep{yoo2022detection, pruthi-etal-2019-combating} have both been studied in natural language processing (NLP) previously, none has studied how to detect OOD and Adv samples when they are simultaneously presented to a deep text classifier.
In previous works, an OOD detection method can only separate the input samples into two groups, but it is unclear if it separates the input based on whether they are ID samples or not, or whether the inputs are OOD samples or not.
If the method separates the input samples based on whether they are ID samples or not, an Adv sample will be categorized into the non-ID groups, falling into the same group as the OOD ones.
Contrarily, when the method separates the input samples based on whether they are OOD samples or not, an Adv sample will be categorized into the non-OOD groups, falling into the same group as the ID samples.
Considering that a text classifier in real-world use case will encounter different types of anomaly and should act differently for different types toward OOD and Adv samples, it is important for a robust text classifier to separate OOD and Adv samples. 

However, identifying OOD and Adv samples from ID samples cannot be achieved before we develop a more unified and in-depth awareness of their differences.
While both kinds of anomalies hurt the model performance, it is unclear whether Adv samples are just some kinds of OOD samples, or they are fundamentally different.
In this work, we aim to answer the following two questions: 
\begin{enumerate}
    \item How different are OOD and Adv samples?
    \item How can we separate OOD, Adv, and ID samples?
\end{enumerate}
To the best of our knowledge, no prior work has answered either of the above questions.
We perform analyses that compare OOD and Adv samples against ID ones from three perspectives: the input features, the hidden representations in different layers of the model, and the output probability distributions of the classifier.
Based on our observations, we propose a simple yet effective method in two stages to separate ID, OOD, and Adv samples: 
first, separate OOD samples from ID and Adv ones based on their hidden features; next, separate Adv from ID ones by the model's output distribution.
Abundant experiments on various combinations of 3 ID datasets, 4 OOD datasets, and 4 Adv attacks verify the effectiveness of our proposed method.

Our contributions are as follows:
\begin{itemize}
    \item
    We find that the hidden representations of the Adv samples are not abnormal until the last few layers of the model, while OOD samples' hidden representations are highly distinguishable from ID ones starting from the first layer.
    \item We find that the maximum probability of the classifier's prediction is an excellent indicator of Adv samples, and this can help one separate Adv samples from ID samples.
    \item We propose an elegant and successful two-staged detection approach to separate ID, OOD, and Adv samples. 
\end{itemize}

\section{Datasets, Attack Methods, and Setup}
In this work, we scrutinize the differences among OOD and ADV samples and attempt to separate them.
Referring to a sample as an OOD one will need to first define what ID samples are.
We briefly introduce the datasets for ID and OOD samples and the attack methods used for generating Adv samples from ID samples. 

\subsection{In-Distribution Datasets}
\label{sec:ID dataset}
\paragraph{IMDB~\citep{maas-etal-2011-learning}}
IMDB is a movie review dataset for binary sentiment classification that contains long and polar movie reviews.
The goal of the task is to predict a piece of movie review to be a positive one or a negative one.
\paragraph{SST-2~\citep{socher-etal-2013-recursive}}
SST-2 is also a binary sentiment classification dataset containing movie reviews.
Very different from IMDB, SST-2 contains pithy reviews whose lengths are significantly less than the reviews in IMDB.
\paragraph{AG-News~\citep{zhang2015character}}
AG-News is a text classification dataset that involves classifying news articles into 4 classes based on their topics, which are world news, sports news, business news, and science/tech news.
\subsection{Out-of-Distribution Datasets}
\label{sec:OOD dataset}
The statistical difference between ID and OOD datasets may include the lengths of the text, the styles, and the topics.
Whether a dataset is out-of-distribution to another dataset depends on how we define in-distribution and out-of-distribution.
For example, SST-2 can be seen as the OOD dataset of IMDB, since their text lengths are very different.
From another viewpoint, both SST-2 and IMDB involve movie reviews and are datasets for binary sentiment classification; they can be considered as in-distribution with each other.
In our work, we simply consider different datasets to be OOD datasets, as in~\citet{hendrycks17baseline}.
In this case, SST-2 and AG-News are both OOD datasets for IMDB; but their degree of out-of-distribution is different.
Each of the three datasets in Section~\ref{sec:ID dataset} serves as an OOD dataset for others.

\paragraph{Yelp Polarity\citep{zhang2015character}} We also include Yelp Polarity as another OOD dataset for each ID dataset in Section~\ref{sec:ID dataset}.
Yelp-Polarity is a dataset for binary sentiment classification which is composed of restaurant reviews.

\subsection{Adversarial Attack Methods}
\label{sec: adv attack}
We follow ~\citet{yoo2022detection} and use four synonym substitution attack methods: Textfooler~\citep{jin2020bert}, Probability Weighted Word Saliency (\citealt{ren-etal-2019-generating}, PWWS), BERT-based Adversarial Examples (\citealt{garg-ramakrishnan-2020-bae}, BAE), and a variant of Textfooler called TF-adj~\citep{morris-etal-2020-reevaluating}.
All of the four methods generate Adv samples by iteratively substituting words in the benign text with their synonyms to flip the prediction of the trained text classifier.
The previous three methods differ in how they determine which word to be substituted first, and what word should be replaced by.
Textfooler and TF-adj are mostly the same, while TF-adj filters legitimate ADV examples with stricter constraints.

We use the Adv samples provided by a recently proposed benchmark~\citep{yoo2022detection}.
They generate Adv examples using the four attacks on the three ID datasets using different models by TextAttack library~\citep{morris2020textattack}.
In this work, we only use the Adv examples generated by attacking the text classifiers fine-tuned from  BERT~\citep{devlin-etal-2019-bert} and RoBERTa~\citep{liu2019roberta}.
Each ID dataset will have four adversarial datasets that are obtained by attacking some ID samples using the four attacks introduced previously.

More details on the datasets and adversarial attacks can be found in Appendix~\ref{app: Datasets and Adversarial Attacks}.

\subsection{Experiment Setup}
\label{subsec: Experiment Setup}
For an ID dataset in Section~\ref{sec:ID dataset}, there will be 3 kinds of OOD datasets (from Section~\ref{sec:OOD dataset}) and 4 kinds of attacks (from Section~\ref{sec: adv attack}), making a total of $12$ different combinations of \{OOD, Adv\}.
We use $\mathcal{D}_{ID}$ and $\mathcal{D}_{OOD}$ to denote ID and OOD dataset.
Specifically, the dataset used to train the text classifier is denoted by $\mathcal{D}_{ID,train}$, while those samples not used to train the classifier form $\mathcal{D}_{ID,test}$.
$\mathcal{D}_{Adv}$ is used to denote the Adv datasets obtained by attacking some ID samples using an adversarial attack method.
To compare the samples from OOD and Adv datasets, we sample $N$ instances from $\mathcal{D}_{ID,test}$, $N$ instances from $\mathcal{D}_{OOD}$, and $N$ instances from $\mathcal{D}_{Adv}$.
We then extract the three types of features (input features, hidden representations, and output probabilities) from the text classifier trained on $\mathcal{D}_{ID,train}$ for analyses and experiments in Section~\ref{sec: analysis} and Section~\ref{sec: separate}.
In all our experiments in Section~\ref{sec: analysis} and Section~\ref{sec: separate}, we set $N$ to $500$.\footnote{\label{footnote: tf-adj} Except SST-2 and AG-News attacked under TF-adj; the generated adversarial examples are far less 500 due to its low attack success rate.
Thus, the results for these two datasets attacked by TF-adj may seem quite different to other datasets and attack methods.
Refer to Table~\ref{tab:N} for details.}
We find that our results are invariant to how we sample those instances from the dataset.

The instances in $\mathcal{D}_{Adv}$ are not generated by attacking those instances in $\mathcal{D}_{ID,test}$ or $\mathcal{D}_{ID,train}$
None of the adversarial examples in $\mathcal{D}_{Adv}$ have benign counterparts in $\mathcal{D}_{ID,test}$ or $\mathcal{D}_{ID,train}$ that only differ in some synonyms.
The process of constructing $\mathcal{D}_{Adv}$ and $\mathcal{D}_{ID,test}$ is detailed in Appendix~\ref{app:Adversarial Samples}.

The text classification models we use are fine-tuned from BERT and RoBERTa provided by TextAttack~\citep{morris2020textattack} and available at Huggingface \href{https://huggingface.co/models}{models}.
We use $C$ to denote the number of classes of the text classifier, for example, $C=2$ for the model trained on SST-2.
We only adopt text classifiers fine-tuned from pre-trained transformer masked language models (MLMs) as their performance are exceptional and widely used in the current NLP community. They are also shown to be better detectors for OOD samples~\citep{hendrycks-etal-2020-pretrained}.

\section{How Different OOD and Adv Samples?}
\begin{wrapfigure}[31]{r}{.45\textwidth}
    \begin{minipage}{\linewidth}
    \centering\captionsetup[subfigure]{justification=centering}
    \includegraphics[trim=60 30 70 50, clip, width=0.95\textwidth]{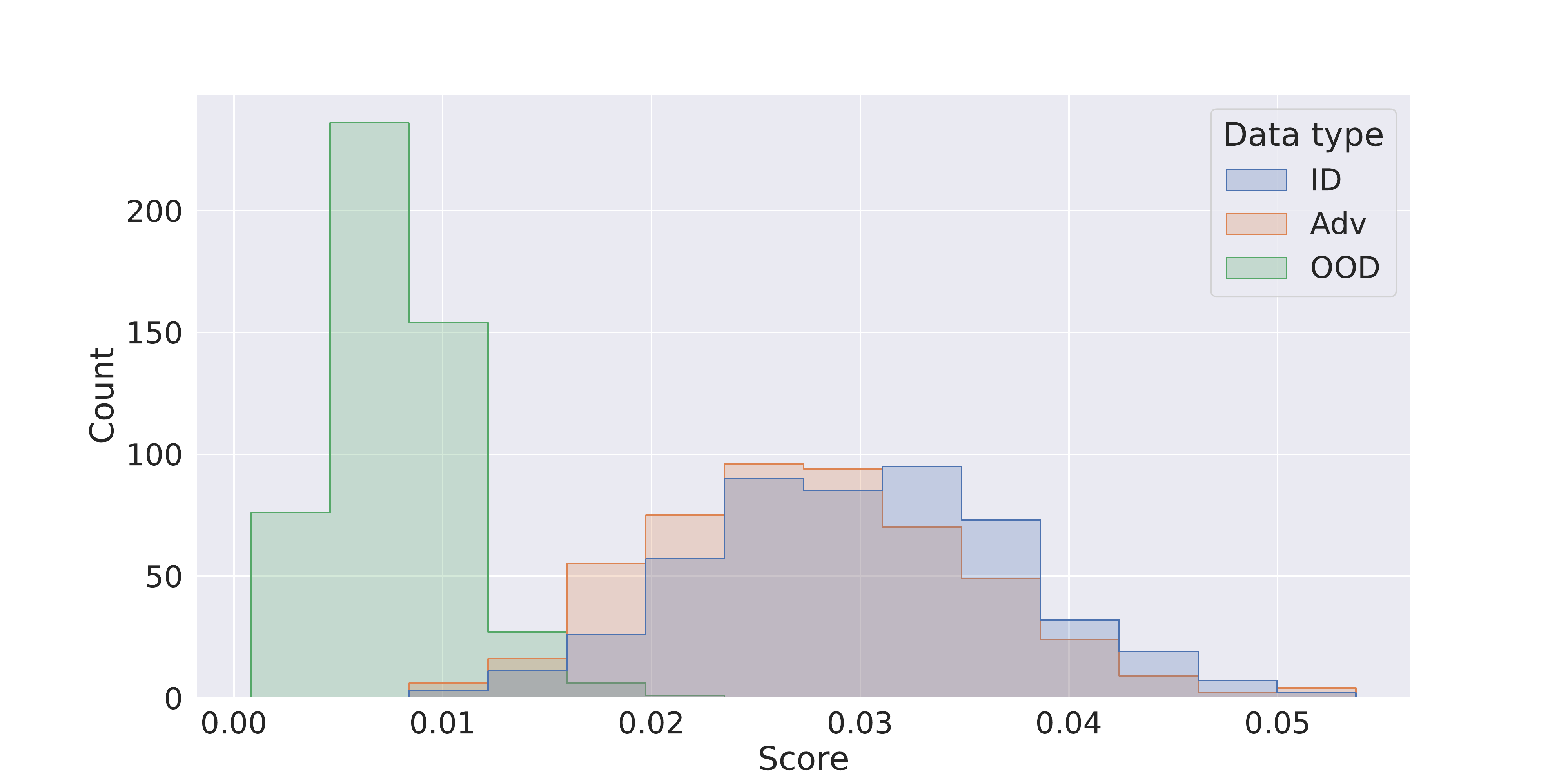}
    \subcaption{$\mathcal{D}_{ID,test}$: IMDB, $\mathcal{D}_{OOD}$: SST-2, $\mathcal{D}_{Adv}$: PWWS.}
    \label{fig:tfidf-mean-bert-imdb-pwws-sst2}
    \includegraphics[trim=60 30 70 50, clip, width=0.95\textwidth]{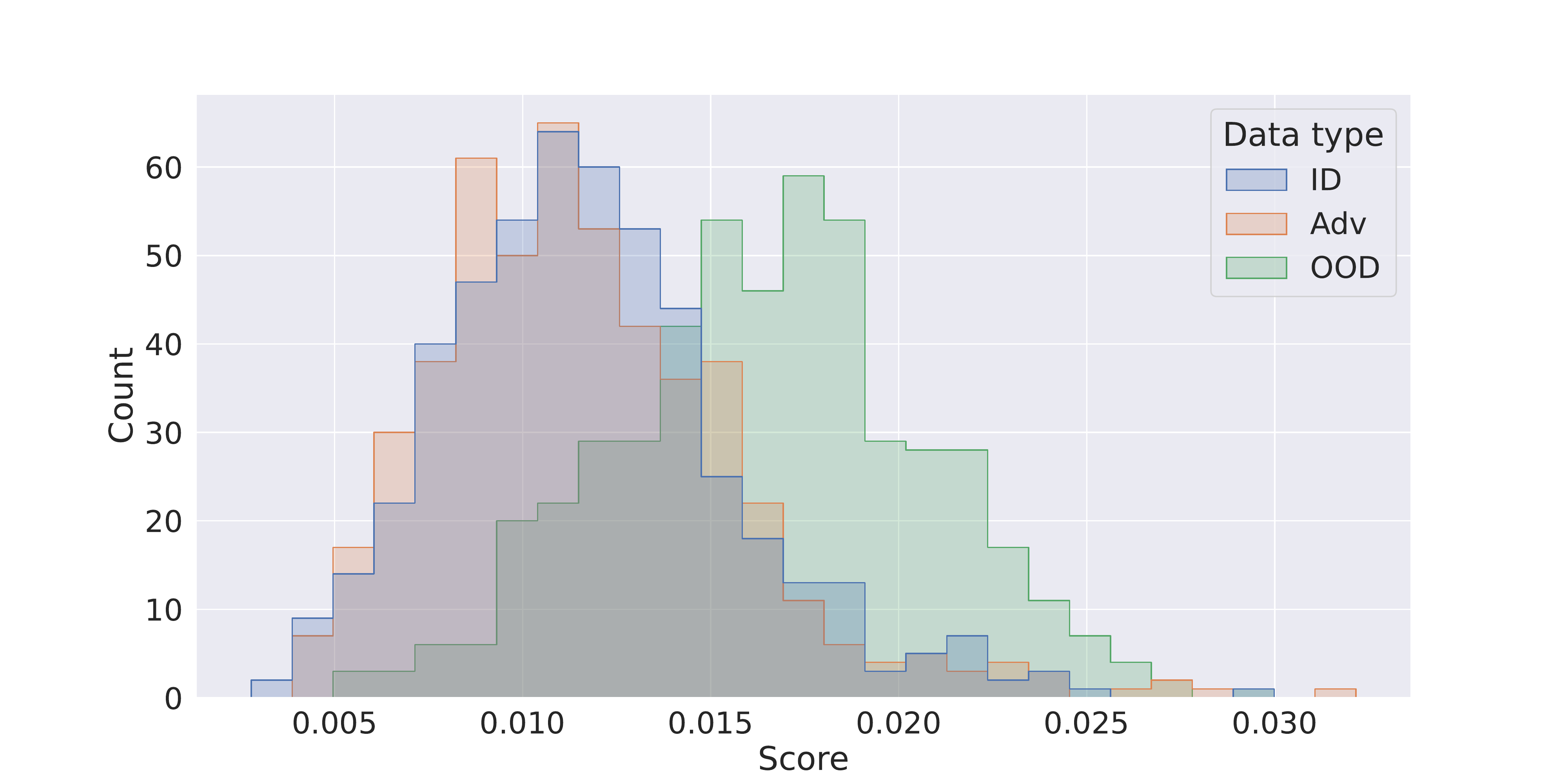}
    \subcaption{$\mathcal{D}_{ID,test}$: AG-News, $\mathcal{D}_{OOD}$: IMDB, $\mathcal{D}_{Adv}$: BAE.}
    \label{fig:tfidf-mean-bert-ag-bae-imdb}
    \includegraphics[trim=60 30 70 50, clip, width=0.95\textwidth]{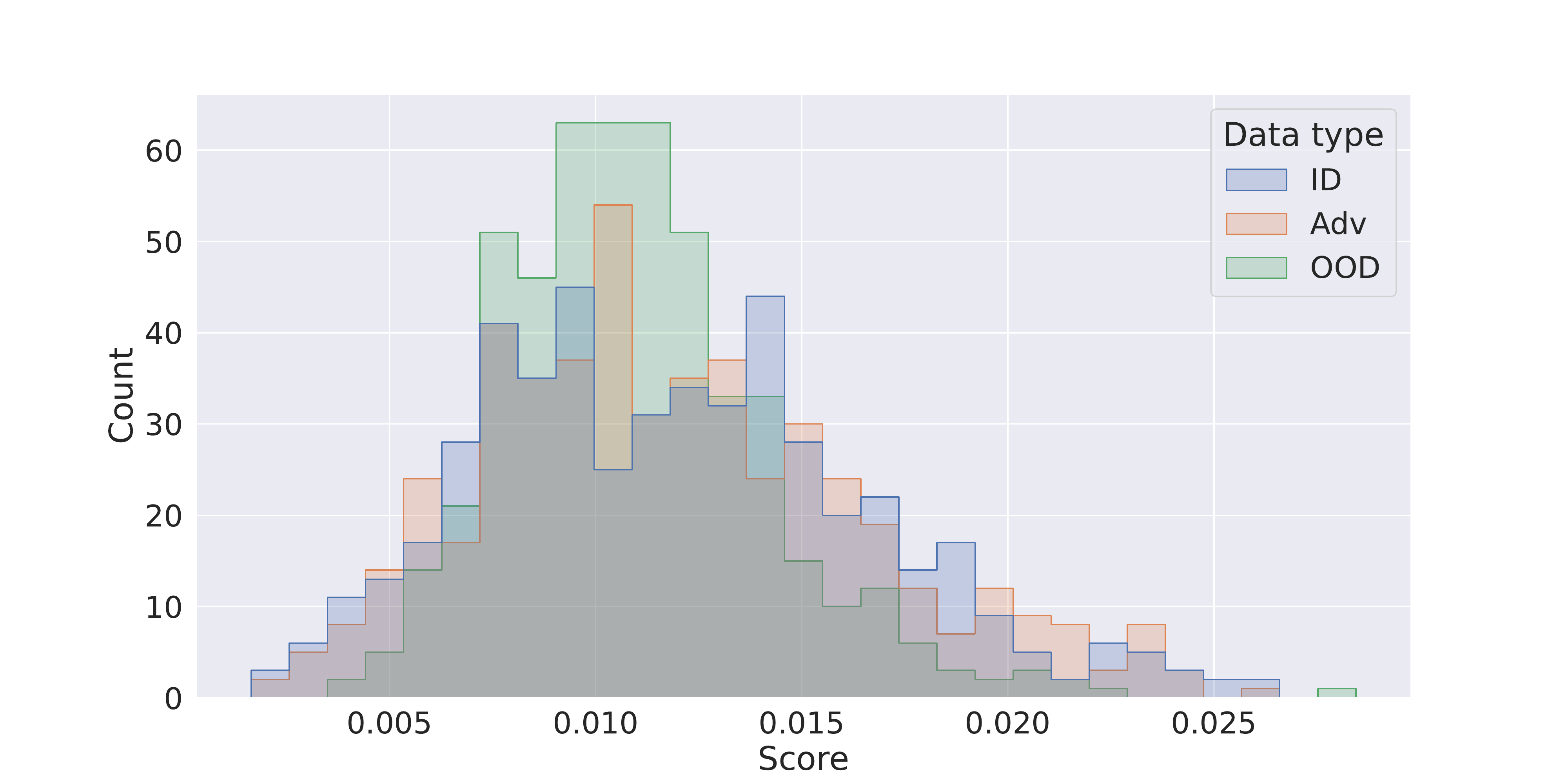}
    \subcaption{$\mathcal{D}_{ID,test}$: SST-2, $\mathcal{D}_{OOD}$: AG-News, $\mathcal{D}_{Adv}$: BAE.}
    \label{fig:tfidf-mean-bert-sst2-bae-ag}
\end{minipage}
\caption{The cosine similarity score between $\mathcal{D}_{ID, train}$ for samples from $\mathcal{D}_{ID,test}$, $\mathcal{D}_{OOD}$, and $\mathcal{D}_{Adv}$.}
\label{fig:tfidf distribution}
\end{wrapfigure}
\label{sec: analysis}
To separate OOD and Adv samples, we first try to understand whether there exists any differences between the two types of data from three aspects: input features, hidden representations, and the output probability distribution of the text classifier.
Additionally, we want to know how abnormal the samples in $\mathcal{D}_{OOD}$ and $\mathcal{D}_{Adv}$ are compared to the in-distribution samples; hence, we also include instances from $\mathcal{D}_{ID,test}$ for the analyses in this section.
This is important since we are not just going to separate OOD and Adv samples, we are separating them \textit{from the ID samples}; thus, we need to understand how they deviate from ID samples.

\subsection{Input Features}
\label{sec: input features}
\subsubsection{Method}

The input features to the text classifier are the words/tokens in the text.
We are interested in knowing whether the words used in OOD and Adv samples are different or not.
We extract the bag-of-word (BOW) feature of each sentence sampled from $\mathcal{D}_{ID,test}$, $\mathcal{D}_{OOD}$, and $\mathcal{D}_{Adv}$ dataset.
The BOW feature is a vector having the dimension of the vocabulary size, and each entry is the TF-IDF of the vocabulary.
We use two different sets of vocabulary for tokenizing the sentences: the tokenizer of bert-base-uncased and roberta-base.

We extract the BOW features using the following procedure. 
First, we sample $8N$ instances for each $C$ classes of $\mathcal{D}_{ID, train}$, and use them to calculate the inverse document frequency (IDF) of each token.
After obtaining the IDFs, we can calculate the BOW features for any given sentence by counting the term frequency (TF) of each token in the sentence and weighted by the IDF of each token.
Last, we normalize the BOW feature such that its $l_2$-norm is $1$.

As an indicator of how different samples drawn from $\mathcal{D}_{ID,test}$, $\mathcal{D}_{OOD}$, or $\mathcal{D}_{Adv}$ are, we compare their cosine similarity score with $\mathcal{D}_{ID, train}$.
For a sentence and its BOW feature, we define its cosine similarity score with $\mathcal{D}_{ID, train}$ as the BOW feature's cosine similarity with the mean of all BOW features of sentences in $\mathcal{D}_{ID, train}$.

\subsubsection{Results}
We plot the distributions of the cosine similarity scores for different combinations of ID, OOD, and Adv datasets, and we select some representative results.
We only show the results when the vocabulary is that of bert-base-uncased as using roberta-base's vocabulary does not change our observations.
For most combinations of $\mathcal{D}_{ID,test}$, $\mathcal{D}_{OOD}$, and $\mathcal{D}_{Adv}$, the results look like Figure~\ref{fig:tfidf-mean-bert-imdb-pwws-sst2}.
The cosine similarity scores of the instances from $\mathcal{D}_{OOD}$ are much lower, compared with $\mathcal{D}_{ID,test}$ and $\mathcal{D}_{Adv}$.
This is understandable since the samples from $\mathcal{D}_{OOD}$ can have a very different vocabulary distribution from $\mathcal{D}_{ID, train}$ if the domains of ID and OOD are different enough.
The cosine similarity scores of $\mathcal{D}_{Adv}$ and $\mathcal{D}_{ID,test}$ behave more similarly, this may spring from the fact that $\mathcal{D}_{Adv}$ are generated from sentences in the in-distribution dataset by swapping less than 25\% of words with their synonyms, and the resulting adversarial examples still have similar BOW features with the in-distribution dataset.

But the previous observations do not always hold: 
In Figure~\ref{fig:tfidf-mean-bert-ag-bae-imdb}, while the distribution of $\mathcal{D}_{Adv}$ and $\mathcal{D}_{ID,test}$ is still very alike, we observe that the BOW features of $\mathcal{D}_{OOD}$ can be even closer with  
$\mathcal{D}_{ID, train}$ compared with $\mathcal{D}_{Adv}$ and $\mathcal{D}_{ID, test}$.
We also find that, in certain combinations of ID and OOD datasets, the results will be like Figure~\ref{fig:tfidf-mean-bert-sst2-bae-ag}: the distribution of the cosine similarity scores of samples from $\mathcal{D}_{ID,test}$, $\mathcal{D}_{OOD}$ and $\mathcal{D}_{Adv}$ are very similar and indistinguishable.

Based on the above results, while the cosine similarity scores of samples from ID, OOD, and Adv datasets can sometimes be very different, using input features to separate ID, OOD, and Adv samples is undoubtedly a bad idea.
This is because we are not able to separate ID and Adv samples simply based on a sentence's BOW feature for all cases in Figure~\ref{fig:tfidf distribution}.
And if we set a threshold of the similarity score, we are not able to ensure whether samples higher than the threshold or lower than the threshold is from $\mathcal{D}_{OOD}$, as in Figure~\ref{fig:tfidf-mean-bert-imdb-pwws-sst2},\ref{fig:tfidf-mean-bert-ag-bae-imdb}.

\subsection{Hidden Representations}
\label{subsec:hidden representations}
\subsubsection{Method}
Next, we aim to understand whether the hidden representations extracted by the model show any differences when the input is from $\mathcal{D}_{OOD}$ and $\mathcal{D}_{Adv}$.
Given the $i^{th}$ sentence $\mathbf{x}_{i}$ from a dataset, each transformer layer $l$ in the text classifier will calculate the hidden representations when forwarding $\mathbf{x}_i$ through the model.
We are interested in how the hidden representations vary from layer to layer, and whether different types of samples will behave differently.

To that end, we fit a maximum likelihood estimator (MLE) for each layer's hidden representations using $\mathcal{D}_{ID, train}$.
We sample $8NC$ samples from $\mathcal{D}_{ID, train}$, each class with $8N$ samples. 
The hidden representations $\mathbf{h}_{i}^{l}$ for each $\mathbf{x}_i$ in $\mathcal{D}_{ID, train}$ through all layers $l\in[1,2,\cdots,12]$ are extracted.
Layer $1$ is the first transformer layer and layer $12$ is the last layer, which is the closest layer to the output. 
$\mathbf{h}_{i}^{l}$ is a tensor of the shape $[T_{i}, d]$, where $T_{i}$ is the number of tokens in $\mathbf{x}_i$ including CLS and SEP tokens, and $d$ is the dimension of a hidden representation for a single token.
To form the hidden representation for fitting the MLE, we aggregate the hidden features of a sentence in a single layer by either 1) taking the hidden representation of the CLS token, denoted as $\mathbf{h}_{i,CLS}^{l}$, or 2) averaging $\mathbf{h}_{i}^{l}$ along the sequence length $T_{i}$, denoted as $\mathbf{h}_{i,Avg}^{l}$.
We will use $\bar{\mathbf{h}}_{i}^{l}$ to refer to either $\mathbf{h}_{i,CLS}^{l}$ or $\mathbf{h}_{i,Avg}^{l}$ later on.
We use $\bar{\mathbf{h}}_{i}^{l}$ to fit a $C$ class-conditional Gaussian distributions with a tied covariance $\mathbf{\Sigma}^{l}$.
The class mean $\mathbf{\mu}_{c}^{l}$ of class $c$ for the multivariate Gaussian and covariance $\mathbf{\Sigma}^{l}$ are estimated by:
\begin{align}
    \hat{\mathbf{\mu}}_c^{l} &= \frac{1}{8N}\sum_{i:y_i=c}\bar{\mathbf{h}}_{i}^{l}, \\
    \hat{\mathbf{\Sigma}}^{l}&=\frac{1}{8NC}\sum_{c}\sum_{i:y_i=c}\big(\bar{\mathbf{h}}_{i}^{l} -\hat{\mathbf{\mu}}_c^{l} \big)\big(\bar{\mathbf{h}}_{i}^{l} -\hat{\mathbf{\mu}}_c^{l} \big)^{\top},
\end{align}
where $y_i\in[1,\cdots,C]$ is the label to $\mathbf{x}_i$. 

After having the MLE parameters fitted, we can assign a score $S^{l}(\mathbf{x}_{i})$ to describe how close a testing sample $\mathbf{x}_{i}$ is to $\mathcal{D}_{ID, train}$ in terms of the hidden features at the $l^{th}$ layer by using the largest log-likelihood of the class-conditional Gaussian distribution, i.e., 
\begin{multline}
\label{eq:score}
    S^l(\mathbf{x}_{i})=\max_{c}\frac{-1}{2}\bigg[
    -\log\frac{1}{{(2\pi)^d|\hat{\mathbf{\Sigma}}^{l}|}} +
    \big(\bar{\mathbf{h}}_{i}^{l} -\hat{\mu}_c^{l} \big)^{\top}(\hat{\mathbf{\Sigma}}^{l})^{-1} \big(\bar{\mathbf{h}}_{i}^{l} -\hat{\mu}_c^{l} \big)
    \bigg],
\end{multline}
where $\bar{\mathbf{h}}_i^{l}$ is the hidden representation for a testing sample $\mathbf{x}_i$ at the $l$-th layer.

\subsubsection{Results}
\begin{wrapfigure}[34]{r}{.45\textwidth}
    \begin{minipage}{\linewidth}
    \centering\captionsetup[subfigure]{justification=centering}
    \includegraphics[trim=60 30 70 50, clip, width=0.95\textwidth]{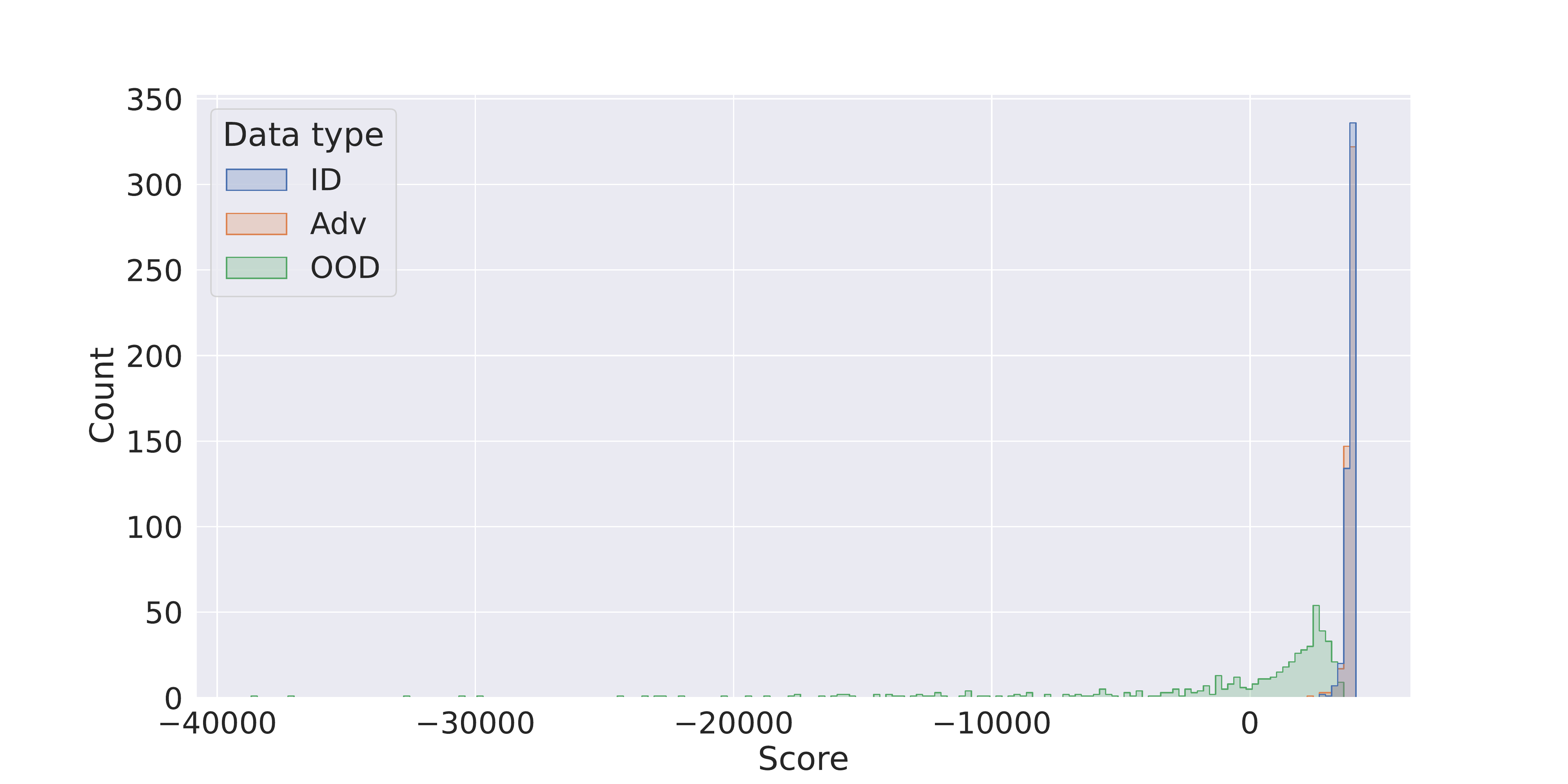}
    \subcaption{Layer 1}
    \label{fig:cls-layer-1-bert-imdb-tf-sst2}\par\vfill
    \includegraphics[trim=60 30 70 50, clip, width=0.95\textwidth]{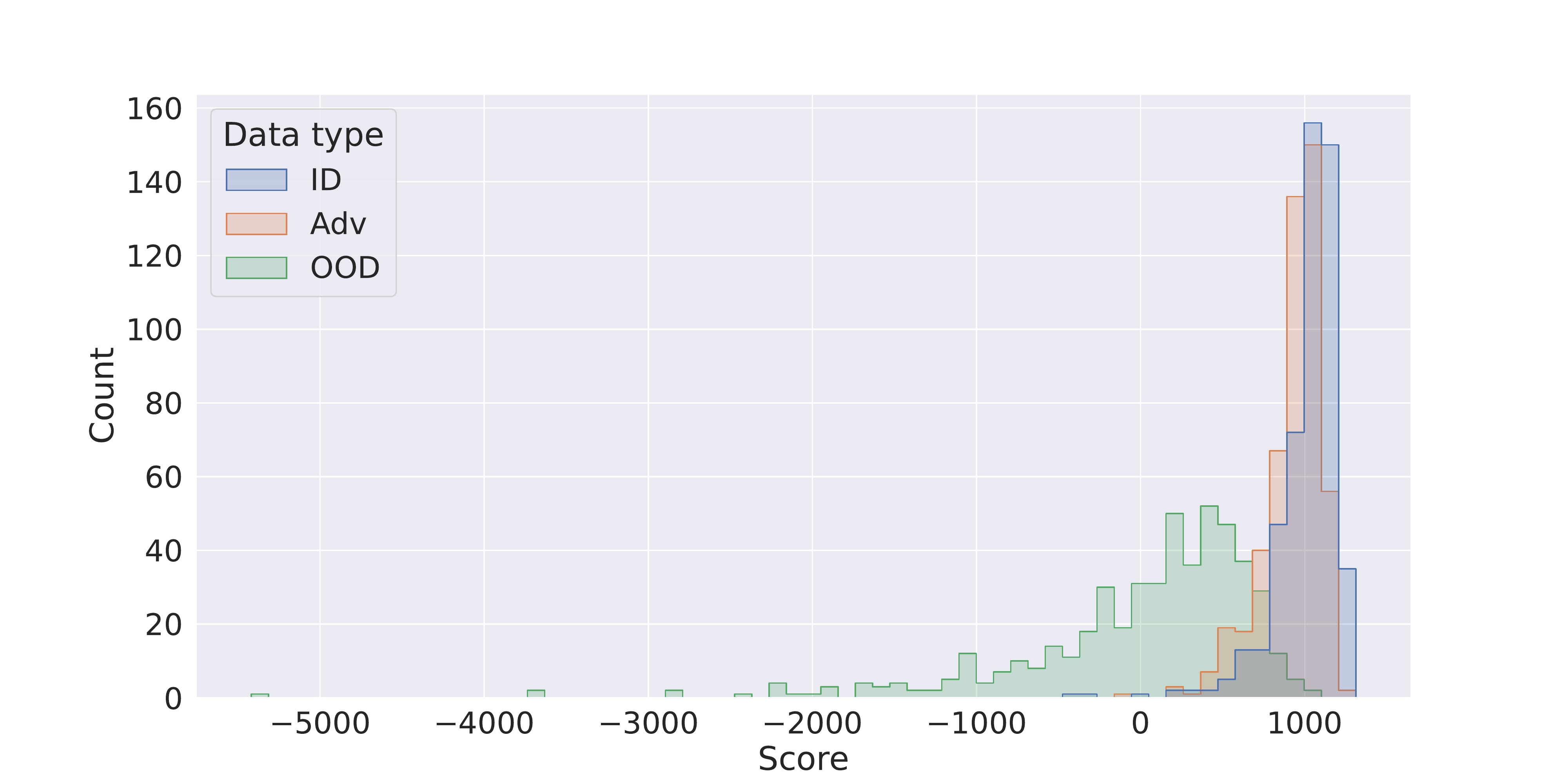}
    \subcaption{Layer 10}
    \label{fig:cls-layer-10-bert-imdb-tf-sst2}
    \includegraphics[trim=60 10 70 50, clip, width=0.95\textwidth]{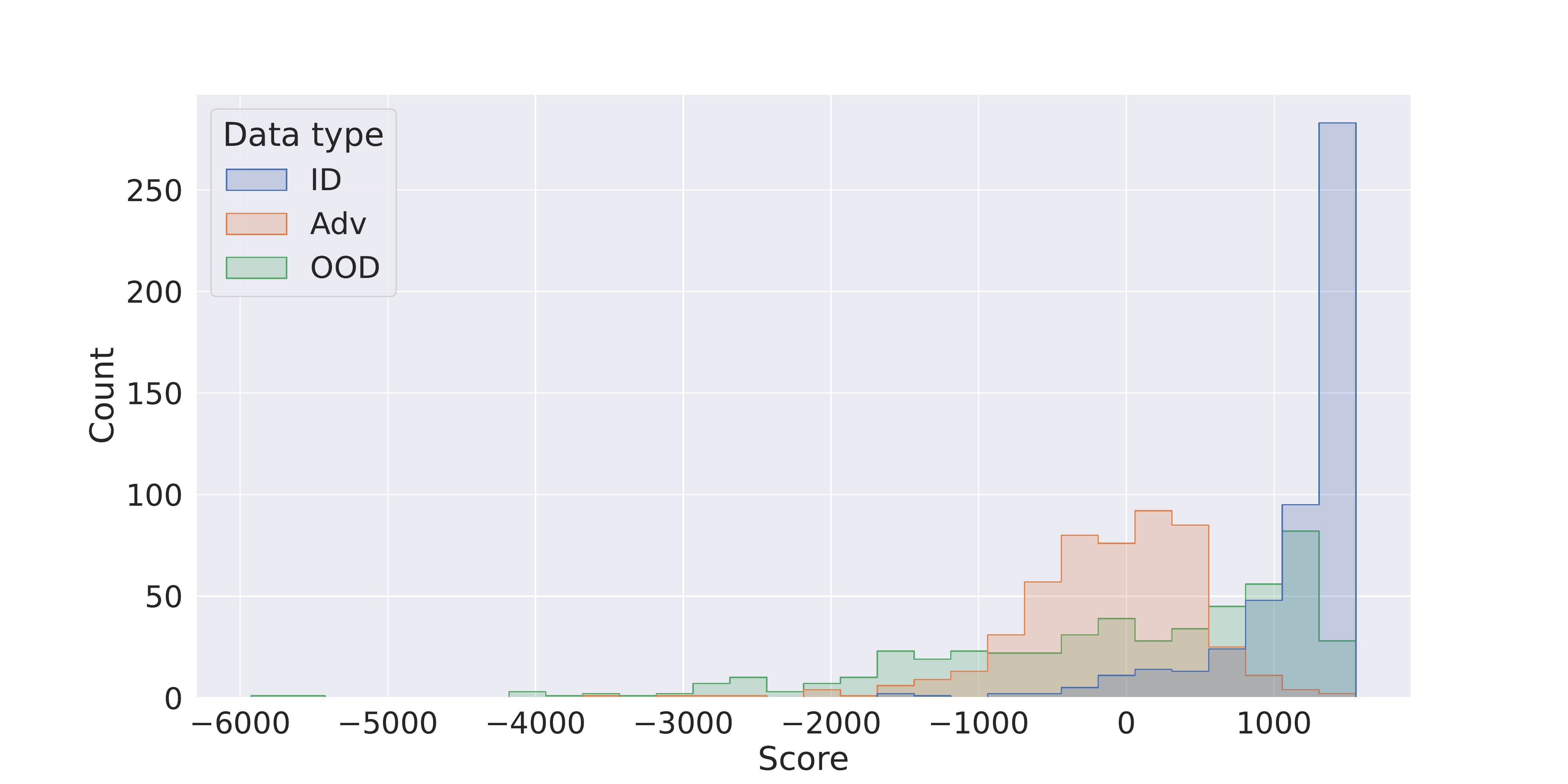}
    \subcaption{Layer 12}
    \label{fig:cls-layer-12-bert-imdb-tf-sst2}
\end{minipage}
\caption{The score $S^{l}(\mathbf{x}_{i})$ of $\mathcal{D}_{ID,test}$, $\mathcal{D}_{OOD}$, and $\mathcal{D}_{Adv}$, the hidden representations among a sentence is aggregated with $\mathbf{h}_{i,CLS}^{l}$.
$\mathcal{D}_{ID,test}$: IMDB, $\mathcal{D}_{OOD}$: SST-2, $\mathcal{D}_{Adv}$: TextFooler. }
\label{fig:mle distribution}
\end{wrapfigure}

We show the distribution of the log likelihood scores calculated using Equation~\ref{eq:score} for samples from $\mathcal{D}_{ID,test}$, $\mathcal{D}_{OOD}$, $\mathcal{D}_{Adv}$ across different layers in Figure~\ref{fig:mle distribution}; the aggregation method used is CLS.
We immediately observe an interesting trend that is very different between $\mathcal{D}_{OOD}$, $\mathcal{D}_{Adv}$.
The $S^{l}(\mathbf{x}_{i})$ distribution of OOD samples is distributed more leftward from ID samples in layer 1 and layer 10, indicating that through those layers, OOD samples are utterly different from ID samples.
As shown in Figure~\ref{fig:cls-layer-12-bert-imdb-tf-sst2}, the hidden representations of OOD samples become much more similar with ID samples in terms of the distribution of the scores, compared with the OOD features in the shallow layers.

Unlike OOD samples, Adv samples' hidden representations much resemble those of ID samples in the first layer.
The abnormality of Adv samples only start to expose deep down in the $10^{th}$ layer, while the score distribution of hidden representations of ID and Adv samples are still highly overlapped.
It was not until the last two layers\footnote{We only show the last layer in Figure~\ref{fig:mle distribution}} that Adv samples can be better distinguished from ID samples.
Instead of gradually deviating from ID samples from the first layer, the anomaly of the hidden representations from Adv samples reveals itself abruptly in the deeper layers of the network.
While we only show a specific combination of $\mathcal{D}_{ID,test}$, $\mathcal{D}_{OOD}$, $\mathcal{D}_{Adv}$, we find that the above observations are very general and hold in all combinations of $\mathcal{D}_{ID,test}$, $\mathcal{D}_{OOD}$, $\mathcal{D}_{Adv}$ we use.
We leave the results for other datasets and adversaries combinations in Appendix~\ref{app: hidden representations}. 
We also find that the above phenomenons can be observed when the hidden representations are $\mathbf{h}_{i,CLS}^{l}$ or $\mathbf{h}_{i,Avg}^{l}$, and they are more pronounced when we use $\mathbf{h}_{i,CLS}^{l}$.

As another way to illustrate how differently the hidden representations from OOD and Adv samples evolve across layers, we design the following experiment.
We evaluate how difficult it is to separate OOD samples from ID samples using the hidden representations of each layer with a threshold-based detector, and we compare it to how difficult it is to separate Adv samples from ID samples with another detector.
A threshold-based detector measures the confidence score of a given sentence, and assigns it as positive if the confidence score is higher than the threshold.
In this subsection, the confidence score is  $S^{l}(\mathbf{x}_{i})$.
Note that in this setting, a detector will be presented with the ID data and \textbf{only one} type of anomaly, either statistical or adversarial, and its job is to determine whether the input is abnormal or not based on the score.
We assign the ID samples as positive (having label 1) and abnormal ones as negative (having label 0).
If the detector can distinguish the abnormal ones from the ID ones, that indicates the two types of samples are distributed far enough for the detector to make the decision.
We report the area under the receiver operating characteristic curve (AUROC) as the performance indicator of the detector.
A perfect detector will have an AUROC of 1 and a random detector will have an AUROC of 0.5.

\begin{wrapfigure}[17]{r}{0.42\textwidth}
  \begin{center}
    \includegraphics[trim=70 0 70 10, clip, width=1.0\linewidth] {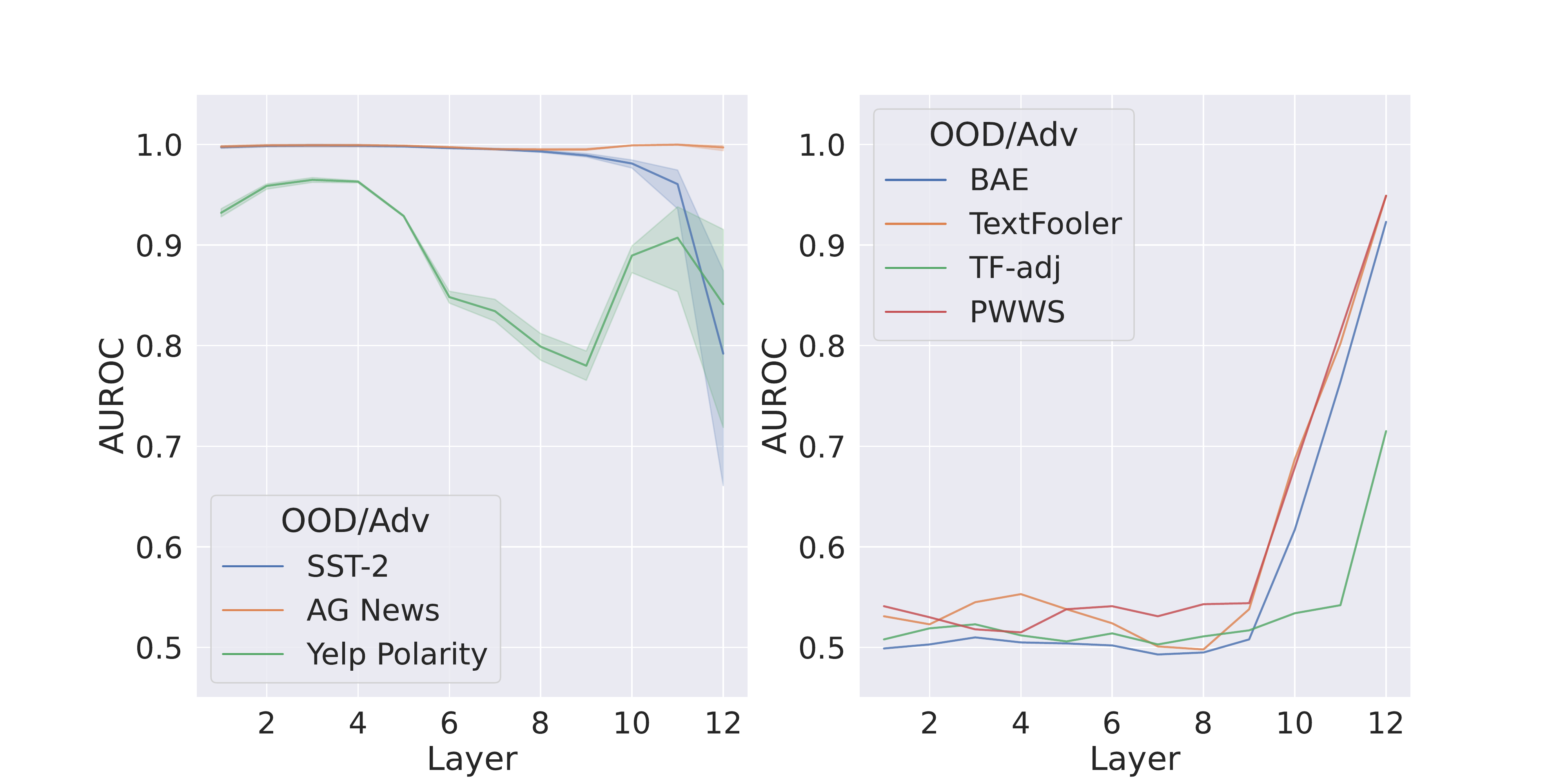}
  \end{center}
  \caption{Left: The detection results to separate $\mathcal{D}_{OOD}$ from $\mathcal{D}_{ID,test}$.
Right: The detection results to separate $\mathcal{D}_{Adv}$ from $\mathcal{D}_{ID,test}$.
$\mathcal{D}_{ID,test}$: IMDB, model: BERT fine-tuned on IMDB, aggregation method: $\mathbf{h}_{i,CLS}^{l}$.
}
\label{fig:cls-bert-imdb-aucroc-2-types}
\end{wrapfigure}
We demonstrate a canonical result when $\mathcal{D}_{ID,test}$ is IMDB in Figure~\ref{fig:cls-bert-imdb-aucroc-2-types}.
First, from the left-hand side, we can easily observe that all the hidden features of three different $\mathcal{D}_{OOD}$ can be easily separated from the features of $\mathcal{D}_{ID,test}$ in early layers.
As the layer gets deeper, the OOD features may get more similar with the ID samples' features, such as SST-2 or Yelp Polarity in Figure~\ref{fig:cls-bert-imdb-aucroc-2-types}, or they can remain dissimilar from ID samples' features, which is the case for AG-News in Figure~\ref{fig:cls-bert-imdb-aucroc-2-types}.

We then turn our attention to the detection results for separating Adv from ID in the right-hand side in Figure~\ref{fig:cls-bert-imdb-aucroc-2-types}.
We see that for all adversarial attacks used for generating the Adv samples, the detection performance is miserable if we use the features from the early layers of the text classifier.
However, the situation twists in the later layers of the network, with a burst in AUROC starting in layer 10 and lasting until the last layer for most Adv samples. 

Again, despite that we only show the results of when $\mathcal{D}_{ID,test}$ is IMDB in Figure~\ref{fig:cls-bert-imdb-aucroc-2-types}, we find all the above observations hold for the other two ID datasets: 
OOD samples can be easily distinguished from ID by their hidden representations starting from the first layer, and they may stay equally identifiable as they get deeper or become more similar with ID samples' hidden representations.
We empirically find that the features from layer 2 are the best for separating $\mathcal{D}_{ID,test}$ and $\mathcal{D}_{OOD}$ for all combinations of ID and OOD datasets; the worst-case AUROC among all the tested combinations is around 0.85.  
Contrarily, Adv samples are so similar to ID samples in terms of the score distribution of their features such that in shallow layers of the network, the two of them are inseparable.
The features of Adv samples only seem out-of-place within the features of ID samples after the $10^{th}$ layer.
The features from the last layer are usually the best for separating ID and Adv samples.
The AUROC for different combinations of $\mathcal{D}_{ID,test}$ and $\mathcal{D}_{Adv}$ may differ, with TextFooler and PWWS mostly the easiest to separate and having an AUROC of at least 0.8, Adv samples generated by BAE is harder to detect and has an AUROC at least 0.7.

\subsection{Output Probability Distribution}
\label{subsec: output probability distribution}
\subsubsection{Method}
In this section, we aim to understand whether the output probability distribution of the text classifier looks different when the input is OOD or Adv samples.
The output probability distribution of a text classifier is a $C$-dimensional vector with each entry representing the probability score of a class. 
It was previously shown that models tend to be over-confident on ID samples, so the maximum probability might be a good indicator for whether a sample is ID or OOD~\citep{hendrycks17baseline}. 
However, it is unclear how confident the text classifier is when faced with Adv samples and whether this confidence can be an indicator of Adv samples. 

To compare the output probability distribution of the models for $\mathcal{D}_{ID,test}$, $\mathcal{D}_{OOD}$, and $\mathcal{D}_{Adv}$, we sample $N$ samples from each of the three datasets.
We extract the output probability distribution of a sample by taking the softmax of the logit distribution from the model's output to form a probability distribution over $C$ classes, and we use the maximum probability among the classes as an indicator of how confident the model is toward the input sample.
We do not use temperature scaling~\citep{liang2018enhancing} when we calculate the probability distribution from the softmax distribution.
\subsubsection{Results}
\begin{figure}[t!]
\begin{center}

\begin{subfigure}{0.45\linewidth}
\begin{center}
\includegraphics[trim=60 30 70 50, clip, width=0.95\textwidth]{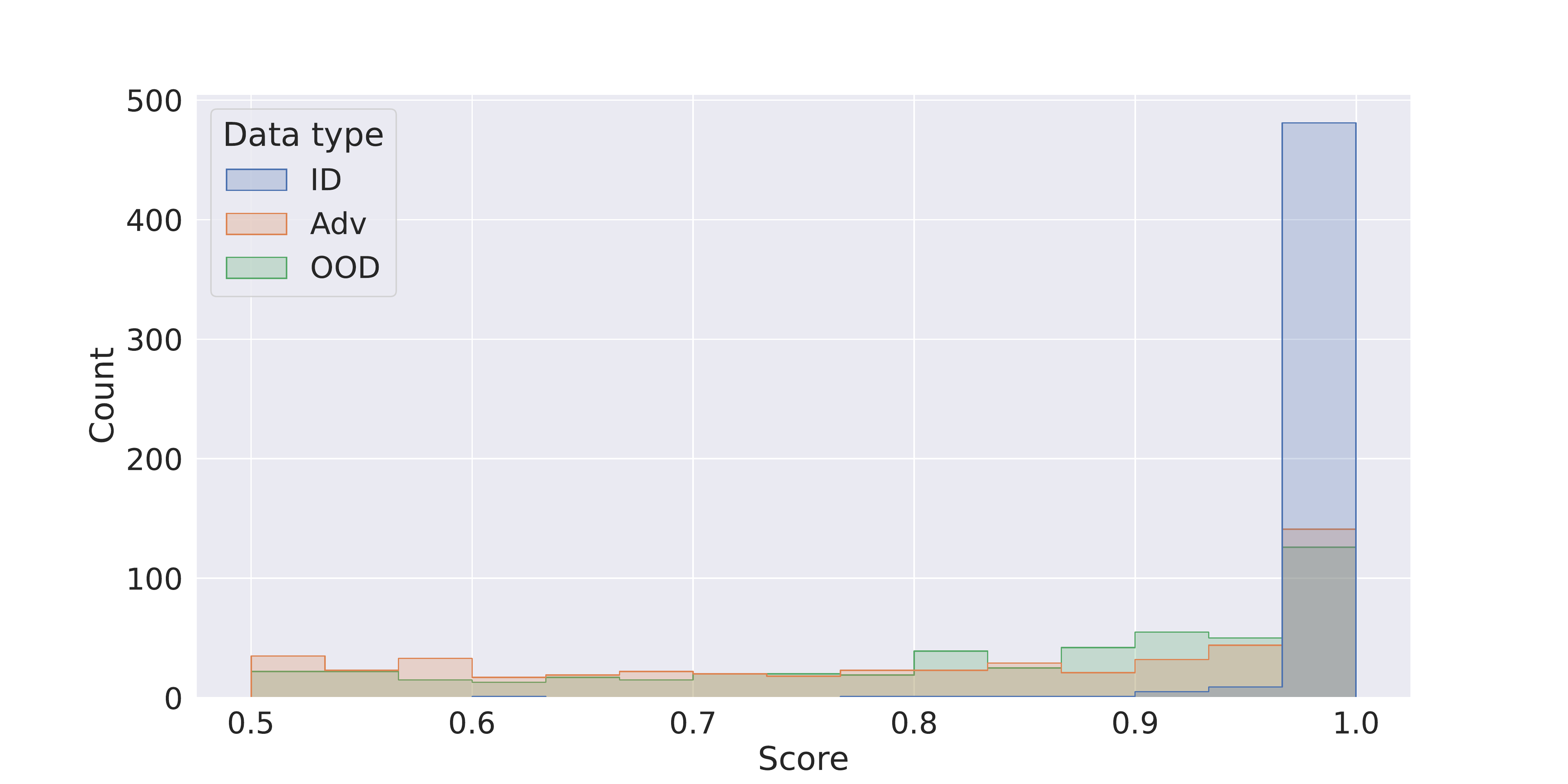}
\caption{$\mathcal{D}_{ID,test}$: IMDB, $\mathcal{D}_{OOD}$: AG-News, $\mathcal{D}_{Adv}$: BAE}
\label{fig:logits-bert-imdb-bae-ag}
\end{center}
\end{subfigure}
\hfill
\begin{subfigure}{0.45\linewidth}
\begin{center}
\includegraphics[trim=60 30 70 50, clip, width=0.95\textwidth]{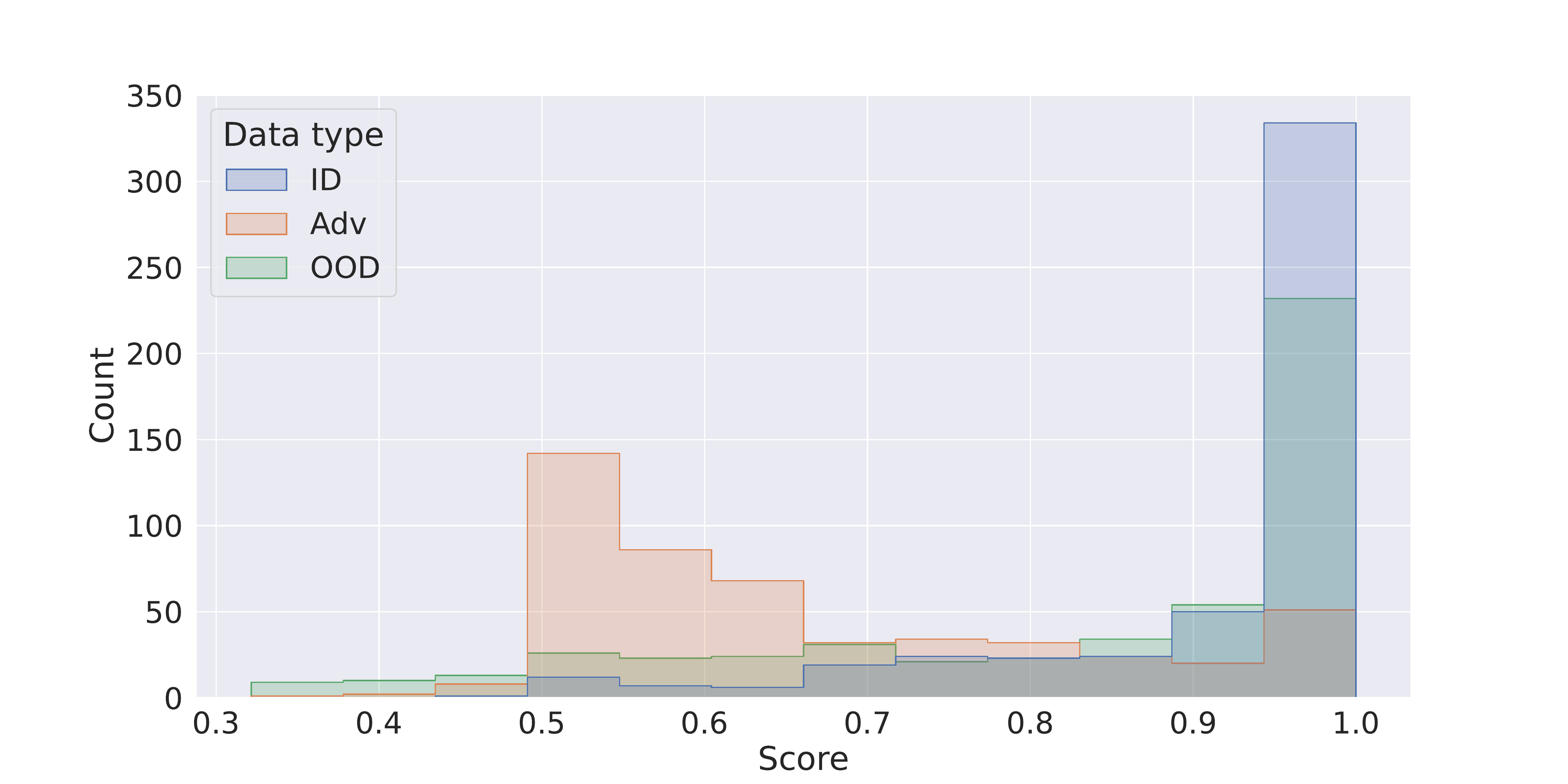}
\caption{$\mathcal{D}_{ID,test}$: AG-News, $\mathcal{D}_{OOD}$: SST-2, $\mathcal{D}_{Adv}$: BAE}
\label{fig:logits-bert-ag-bae-sst2}
\end{center}
\end{subfigure}

\caption{The maximum probabilities of the text classification models fine-tuned from BERT for different combinations of $\mathcal{D}_{ID,test}$, $\mathcal{D}_{OOD}$, and $\mathcal{D}_{Adv}$.
}
\label{fig:logits distribution}
\end{center}
\end{figure}

Overall, we find that there are two different patterns for the distribution of the maximum probability that mainly differ in how the maximum probabilities of samples from $\mathcal{D}_{OOD}$ distribute, as displayed in Figure~\ref{fig:logits-bert-imdb-bae-ag} and ~\ref{fig:logits-bert-ag-bae-sst2}.
For samples from $\mathcal{D}_{ID,test}$, the models are always confident about them, and the maximum probabilities are mostly around 1.
Contrarily, the models are consistently unsure about their prediction when the input is from $\mathcal{D}_{Adv}$, and the maximum probabilities scatter from $1/C$ to $1$.
The maximum probability distribution for samples from OOD datasets is rather conflicting for different combinations of $\mathcal{D}_{ID,test}$ and $\mathcal{D}_{OOD}$.
The model can be quite confident on samples from OOD as in Figure~\ref{fig:logits-bert-ag-bae-sst2}, but it can also be far less assured as in Figure~\ref{fig:logits-bert-imdb-bae-ag}.

The above interesting observations from Figure~\ref{fig:logits distribution} give the following message:
The maximum probability can be lousy at determining whether an input is OOD or not, but it sure has the potential to be used in detecting Adv samples.

\section{Separating OOD and Adv Samples}
\label{sec: separate}
A text classification model needs to be able to distinguish between OOD and Adv samples.
When a text classifier sees an OOD sample, the model cannot handle the input since it is something it has not been trained on.
The model should not predict an OOD sample since the prediction may not even make sense.
On the other hand, when the model is input with an ADV sample, the input can be converted into its benign counterpart for prediction.
Knowing that the current input is an adversarial one also helps the model developer to know that some malicious users may be attacking the model, and measures need to be undertaken to secure the model.
In this section, we focus on designing a method to separate OOD and Adv samples.

Given an input sentence, we want to tell which kind of anomaly it is \textit{if it is indeed an anomaly}.
Thus, the proposed method needs to be able to separate ID, OOD, and Adv samples.
Existing works on OOD detections~\citep{xu-etal-2021-unsupervised, hendrycks-etal-2020-pretrained, xu-etal-2020-deep} do not discuss how their proposed method behaves when the input is Adv sample, and previous works on Adv detection~\citep{yoo2022detection, pruthi-etal-2019-combating} do not test their methods on OOD samples.
~\citet{lee2018simple} proposed a general framework to detect both OOD and ADV examples, but they do not discuss how to separate the two types of anomaly.

The scenario in this section is as follows:
We have a trained text classifier and the corresponding $\mathcal{D}_{ID, train}$, and we want to separate OOD, Adv, and ID samples using some threshold-based detectors, instead of training a classifier to distinguish different types of samples~\citep{zhou-etal-2019-learning}.
The threshold of the detectors can be determined with or without the knowledge of the OOD and Adv datasets, detailed in Appendix~\ref{app: Setting the Thresholds in Stage 1 and Stage 2}.

\subsection{Method}

We propose to detect whether a given input is from $\mathcal{D}_{ID,test}$, $\mathcal{D}_{OOD}$ or $\mathcal{D}_{Adv}$ by the following two-stage pipeline:
In the \textbf{first stage}, we use the hidden representations from the earlier layers to build a threshold-based OOD detector using $S^{l}(\mathbf{x}_{i})$, whose goal is to separate OOD samples from ID and Adv samples.
If the detector thinks that input is not an OOD sample, then we proceed to the second stage.
In the \textbf{second stage}, we use the maximum output probability of the model to build another threshold-based detector that aims to distinguish Adv samples from ID samples.

Using hidden features from shallow layers to isolate OOD samples in stage 1 is feasible based on the following observation in Section~\ref{subsec:hidden representations}: OOD samples deviate from ID samples in the earlier layers, while Adv ones highly resemble ID ones in the earlier layers and only start to deviate from ID samples in the deeper layers.
The detector can thus separate those hidden features into two groups, one that does not resemble ID samples and one that resemble ID samples (including ID and Adv ones).
Using the maximum probability to separate Adv samples from ID samples in stage 2 should work as we observed that models tend to be unconfident when the inputs are Adv samples, as shown in Section~\ref{subsec: output probability distribution}.
While model can also be unconfident about some OOD samples, they will not cause any problems in stage 2 as they should be detected in stage 1 and not proceeding to stage 2.

Our proposed framework is simple and does not require additional models for detection; we only need the class-conditional Gaussian's parameters and the threshold for the two detectors.

\subsection{Experiment}
In the experiment in this section, we sample the instances as stated in Section~\ref{subsec: Experiment Setup}.
In stage 1, we assign the OOD samples as negative, and we assign \textbf{both} ID samples and Adv samples as positive.
This is different from our experiment in Section~\ref{subsec:hidden representations}, in which the detectors will only be presented with a \textbf{single} type of anomaly and the ID samples.
We use the score $S^2(\mathbf{x}_{i})$, the score calculated based on the hidden representations from the second layer, to build the detector in stage 1.

The results are shown in Table~\ref{tab:stage 1}. 
We see that under all combinations of ID, OOD, and Adv datasets, we can build a detector that can almost perfectly separate OOD samples from  ID and Adv samples.
While in Table~\ref{tab:stage 1}, we build a detector for each combination of \{$\mathcal{D}_{ID,test}$, $\mathcal{D}_{OOD}$, $\mathcal{D}_{Adv}$\}, it is possible to use only one detector to separate all different $\mathcal{D}_{OOD}$ from a fixed $\mathcal{D}_{ID,test}$ and all different $\mathcal{D}_{Adv}$.
This can be done easily by selecting the highest threshold among the thresholds of the detectors of different OOD detectors for a fixed $\mathcal{D}_{ID,test}$ (more precisely explained in Appendix~\ref{app: faq}.3).

\begin{table}[t]
\caption{\ref{tab:stage 1}:The AUROC of the detectors for separating OOD samples from ID and Adv samples.
    The AUROC for an entry is the averaged AUROC for a specific combination of \{$\mathcal{D}_{ID,test}$, $\mathcal{D}_{OOD}$\} when varying $\mathcal{D}_{Adv}$.
    The variance across different $\mathcal{D}_{Adv}$ is small and not shown.
    \ref{tab:stage 2}: The AUROCs of the detectors for separating Adv samples from ID ones.
    }
\label{tab:detect full result}
\begin{subtable}[t]{0.48\textwidth}
    \centering
    \begin{tabular}{c|ccc}
        \hline
        \diagbox[width=5em]{OOD}{ID} & IMDB & SST-2 & AG-News\\
        \hline \hline
        IMDB & - & 0.99& 0.93\\
        SST-2 & 1.00& -& 0.86 \\
        AG-News & 1.00 & 0.98 & - \\
        Yelp Polarity &0.96 & 0.97 & 0.87 \\
        \hline 
    \end{tabular}
    \caption{}
    \label{tab:stage 1}
\end{subtable}
\hfill
\begin{subtable}[t]{0.48\textwidth}
    \centering
    \begin{tabular}{c|cccc}
        \hline
        \diagbox[width=5em]{Adv}{ID} & IMDB & SST-2 & AG-News\\
        \hline \hline
        TextFooler & 0.98 & 0.90 & 0.97\\
        PWWS & 0.98 &0.88 & 0.94 \\
        BAE & 0.97 & 0.75 & 0.89 \\
        TF-adj &0.82 & 0.85 & 0.88 \\
        \hline 
    \end{tabular}
    \caption{
    }
    \label{tab:stage 2}
\end{subtable}

\end{table}

Having the OOD samples separated in stage 1, the detector in stage 2 only needs to focus on dividing ID and Adv samples.
We use another threshold-based detector, which makes the decision based on the maximum probability discussed in Section~\ref{subsec: output probability distribution}.
The detector should assign the ID samples as positive and the Adv samples as negative.
In our experiments, we assume that the detector in stage 1 is ideal such that all OOD samples will be detected in stage 1 and not presented to the detector in stage 2.
This is a reasonable assumption, given that the AUROCs of the detectors in stage 1 are very high.

The detection results for stage 2 is presented in Table~\ref{tab:stage 2}.
We see that the detectors in stage 2 give a very decent performance on all combinations of ID and Adv datasets.
Note that the setting in stage 2 is the same as in \citet{yoo2022detection}, where they aim to detect Adv samples among ID samples; our results in Table~\ref{tab:stage 2} show that detection based on the maximum probability is a simpler yet equally powerful detection method.
While we build different detectors for different Adv datasets for a fixed ID dataset, it is also possible to build a single detector for all possible Adv datasets.
This is because we already know the maximum probability of Adv samples is lower for all kinds of Adv datasets. 

Combining the detectors in stage 1 and stage 2, we can separate ID, OOD, and Adv samples.
We leave the experiment setup and results of cascading stage 1 and 2 in Appendix~\ref{app: Experiments for Cascading Stage 1 and Stage 2}.
Simple as our method is, it shows non-trivial results on separating ID, OOD, and Adv samples, an important task that has been overlooked by the NLP community.

\section{Discussion and Conclusion}
\label{sec: Discussion and Conclusion}
In this work, we analyze the differences between two types of anomalies, OOD and Adv samples benchmarked against ID samples.
We conduct comprehensive analyses to characterize OOD and Adv samples from three different aspects: the input features, the hidden representations across different layers of the model, and the output probability distributions of the model.
We show that the similarity score based on BOW features is unstable among different combinations of ID/OOD/Adv data, which refrains us from utilizing them for separating OOD samples and Adv samples.
From the aspect of hidden representations, we observe that OOD samples are very different from ID samples starting from the first layer of the model.
On the contrary, Adv samples seem just like the ID ones for the first nine layers, and their abnormality only becomes evident in the last three layers.
We also find that the models tend to be less confident, in terms of the maximum probability, for Adv samples.
Based on our analysis, we propose an original, simple yet effective two-staged detection method to separate the ID, OOD, and Adv samples.
We show the superiority of our proposed method by abundant experiments over various combinations of ID, OOD, and Adv datasets.
The analyses in our work elevate our knowledge of anomalies that bring down the model, and we believe our proposed method will help build more robust text classification models.

When deploying a deep learning model in real-world, it is important for it to act properly when faced with anomaly examples, and the first step to achieve this goal is to correctly identify it as an anomaly.
Moreover, knowing what kind of anomaly, including Adv and OOD samples, is presented to the model will make the model to process the anomalous data in a more robust way.
While we only focus on separating the Adv and OOD samples in text classification, we believe that this is an important topic for developing more trust-worthy machine learning models in all domains, including computer vision, speech processing, and other non-classification tasks in NLP.
As a pioneering work in this topic, we believe our thorough analysis and exhaustive experiments shed lights on this topic.

\newpage

\bibliography{custom}
\bibliographystyle{natbib}

\section*{Checklist}

The checklist follows the references.  Please
read the checklist guidelines carefully for information on how to answer these
questions.  For each question, change the default \answerTODO{} to \answerYes{},
\answerNo{}, or \answerNA{}.  You are strongly encouraged to include a {\bf
justification to your answer}, either by referencing the appropriate section of
your paper or providing a brief inline description.  For example:
\begin{itemize}
  \item Did you include the license to the code and datasets? \answerYes{See Section~\ref{gen_inst}.}
  \item Did you include the license to the code and datasets? \answerNo{The code and the data are proprietary.}
  \item Did you include the license to the code and datasets? \answerNA{}
\end{itemize}
Please do not modify the questions and only use the provided macros for your
answers.  Note that the Checklist section does not count towards the page
limit.  In your paper, please delete this instructions block and only keep the
Checklist section heading above along with the questions/answers below.

\begin{enumerate}

\item For all authors...
\begin{enumerate}
  \item Do the main claims made in the abstract and introduction accurately reflect the paper's contributions and scope?
    \answerYes{See Abstract\ref{abstract} and \ref{sec: intro}.}
  \item Did you describe the limitations of your work?
    \answerYes{In Section\ref{sec: Discussion and Conclusion}.}
  \item Did you discuss any potential negative societal impacts of your work?
    \answerYes{See Appendix~\ref{app: Broader Impact and Ethical Impact}.}
  \item Have you read the ethics review guidelines and ensured that your paper conforms to them?
    \answerYes{See Appendix~\ref{app: Broader Impact and Ethical Impact}.}
\end{enumerate}

\item If you are including theoretical results...
\begin{enumerate}
  \item Did you state the full set of assumptions of all theoretical results?
    \answerNA{}
        \item Did you include complete proofs of all theoretical results?
    \answerNA{}
\end{enumerate}

\item If you ran experiments...
\begin{enumerate}
  \item Did you include the code, data, and instructions needed to reproduce the main experimental results (either in the supplemental material or as a URL)?
    \answerYes{The code is submitted with the paper and will be made publicly available once the paper is accepted.}
  \item Did you specify all the training details (e.g., data splits, hyperparameters, how they were chosen)?
    \answerYes{See Appendix~\ref{app: Datasets and Adversarial Attacks}.}
        \item Did you report error bars (e.g., with respect to the random seed after running experiments multiple times)?
    \answerYes{We provide some error bars as in Figure~\ref{fig:cls-bert-imdb-aucroc-2-types}.
    As we stated in Section 2.4, we do not report the variations among different runs since their variance is small.}
        \item Did you include the total amount of compute and the type of resources used (e.g., type of GPUs, internal cluster, or cloud provider)?
    \answerYes{See Appendix~\ref{app: Run Time and Infrastructure}.}
\end{enumerate}

\item If you are using existing assets (e.g., code, data, models) or curating/releasing new assets...
\begin{enumerate}
  \item If your work uses existing assets, did you cite the creators?
    \answerYes{See Section~\ref{sec:ID dataset} and Section~\ref{sec:OOD dataset}.}
  \item Did you mention the license of the assets?
    \answerNo{We are not able to find the licenses of the assets.}
  \item Did you include any new assets either in the supplemental material or as a URL?
    \answerNA{}
  \item Did you discuss whether and how consent was obtained from people whose data you're using/curating?
    \answerNo{We cannot find the license of the datasets, but we believe that their releasing of the datasets is for the community to use them.}
  \item Did you discuss whether the data you are using/curating contains personally identifiable information or offensive content?
    \answerYes{See Appendix~\ref{app: Datasets and Adversarial Attacks}.}
\end{enumerate}

\item If you used crowdsourcing or conducted research with human subjects...
\begin{enumerate}
  \item Did you include the full text of instructions given to participants and screenshots, if applicable?
    \answerNA{}
  \item Did you describe any potential participant risks, with links to Institutional Review Board (IRB) approvals, if applicable?
    \answerNA{}
  \item Did you include the estimated hourly wage paid to participants and the total amount spent on participant compensation?
    \answerNA{}
\end{enumerate}

\end{enumerate}


\appendix

\section{Broader Impact and Ethical Impact}
\label{app: Broader Impact and Ethical Impact}
Our work aims toward building more robust text classifiers.
We believe our work has positive contribution on making AI system more trust-worthy.
Comparing OOD and Adv samples in order to build more robust models is not only important for text classification models, but also necessary for all deep learning models in natural language processing, computer vision, and speech processing.
We highlight the importance of being able to separate OOD and Adv samples for a model, which is a critical research problem that has yet to be emphasized.
It is also a real problem a model will need to tackle when it is deployed in our daily life.
We open a new and crucial direction for improving model robustness in the AI community.

Our paper is motivated by making the text classification models more robust.
We cannot think of any possibility that our work may be mis-used by any vicious user with bad intention.
However, if a malicious user knows that the model can detect Adv samples, he or she may try to craft stronger adversaries to bypass the detection method; this is the only possible risk we can think of.
We believe that we do not violate the Ethic Guidelines in NeurIPS 2022.

\section{Datasets and Adversarial Attacks}
\label{app: Datasets and Adversarial Attacks}
\subsection{Datasets}
For all the ID and OOD datasets in Section~\ref{sec:ID dataset} and Section~\ref{sec:OOD dataset}, we load them using the Huggingface Datasets library~\citep{quentin_lhoest_2021_5639822,lhoest-etal-2021-datasets}.
In our experiments, $\mathcal{D}_{ID, train}$ is the training split of the ID dataset, $\mathcal{D}_{OOD}$ is the testing split of OOD the dataset.
How $\mathcal{D}_{ID,test}$ and $\mathcal{D}_{Adv}$ are obtained will be explained in the next subsection.

Datasets we use are mostly for sentiment analysis, which means they may contain subjective opinions that may be offensive to some individuals.
However, we leave them as is because removing those possibly offensive languages may make the sentiment analysis task unable to be conducted.
Also, removing specific kinds of words from the dataset during testing may cause the distribution to be slightly different with the distribution that the text classifier is trained on.
Given that our goal includes separating ID and OOD samples, we do not want to create any possible mismatch in the ID training data and ID testing data.

\subsection{Adversarial Samples}
\label{app:Adversarial Samples}
We download the Adv from \href{https://drive.google.com/file/d/1VMiyg5Mrwwhz-156F4PH7-mTNla_CYJ1/view?usp=sharing}{here} provided by ~\citet{yoo2022detection}.
They provide sentences resulting from attacking the text classifiers using TextAttack~\citep{morris2020textattack}, including successful and failed adversaries, and the attack is done on the testing or development split of the ID dataset.
For an attack result for a specific text classifier using a specific adversarial attack, we first select those instances that successfully fool the text classifier.
For those successful adversaries, we split them into two groups with an equal number of samples, and one of them will be used as the $\mathcal{D}_{Adv}$.
For the other group of successful adversaries, we take their \textbf{benign counterparts} to form the $\mathcal{D}_{ID,test}$.
In case when $\mathcal{D}_{ID,test}$ contains too few samples, we will add the benign counterparts in the failed attacks to enlarge the size of $\mathcal{D}_{ID,test}$.
This way, we can ensure that $\mathcal{D}_{ID,test}$ and $\mathcal{D}_{Adv}$ will not contain pairs of samples that are benign/adversarial to each other.

For input preprocessing, we remove the '<br>', '<br />', and '//' in some datasets to prevent the model from distinguishing ID and OOD samples based on those artifacts.

\section{Implementation Details}
\subsection{Hidden Features}
In Equation~\ref{eq:score}, there is a $(2\pi)^{d}$ term in the first summand.
In our implementation, we omit that $(2\pi)^{d}$ term as it is a constant when $d$ is fixed, which is $768$.
Thus, the score $S^{l}(\mathbf{x}_{i})$ is in fact calculated by the following formula: 
\begin{multline}
\label{eq:score simplified}
    S^l(\mathbf{x}_{i})=\max_{c}\frac{-1}{2}\bigg[
    -\log\frac{1}{{|\hat{\mathbf{\Sigma}}^{l}|}} +
    \big(\mathbf{h}_{i}^{l} -\hat{\mu}_c^{l} \big)^{\top}(\hat{\mathbf{\Sigma}}^{l})^{-1} \big(\mathbf{h}_{i}^{l} -\hat{\mu}_c^{l} \big)
    \bigg].
\end{multline}
The $S^{l}(\mathbf{x}_{i})$ show in Figure~\ref{fig:mle distribution} and Figure~\ref{fig:roberta mle distribution} is calculated using Equation~\ref{eq:score simplified}.
\subsection{Run Time and Infrastructure}
\label{app: Run Time and Infrastructure}
Our codes are implemented with Pytorch, and we use Huggingface Transformers~\citep{wolf-etal-2020-transformers} and datasets~\citep{lhoest-etal-2021-datasets} to load fine-tuned text classification models and datasets.
Running the full set of our experiments takes less than 24 hours with a single Tesla V100.
All of the text classification models have around 110M parameters.
\section{Supplementary Figures for Section~\ref{subsec:hidden representations}}
\label{app: hidden representations}
\begin{figure}[t!]
\begin{subfigure}{0.42\textwidth}
\begin{center}
\includegraphics[trim=60 30 70 50, clip, width=0.95\textwidth]{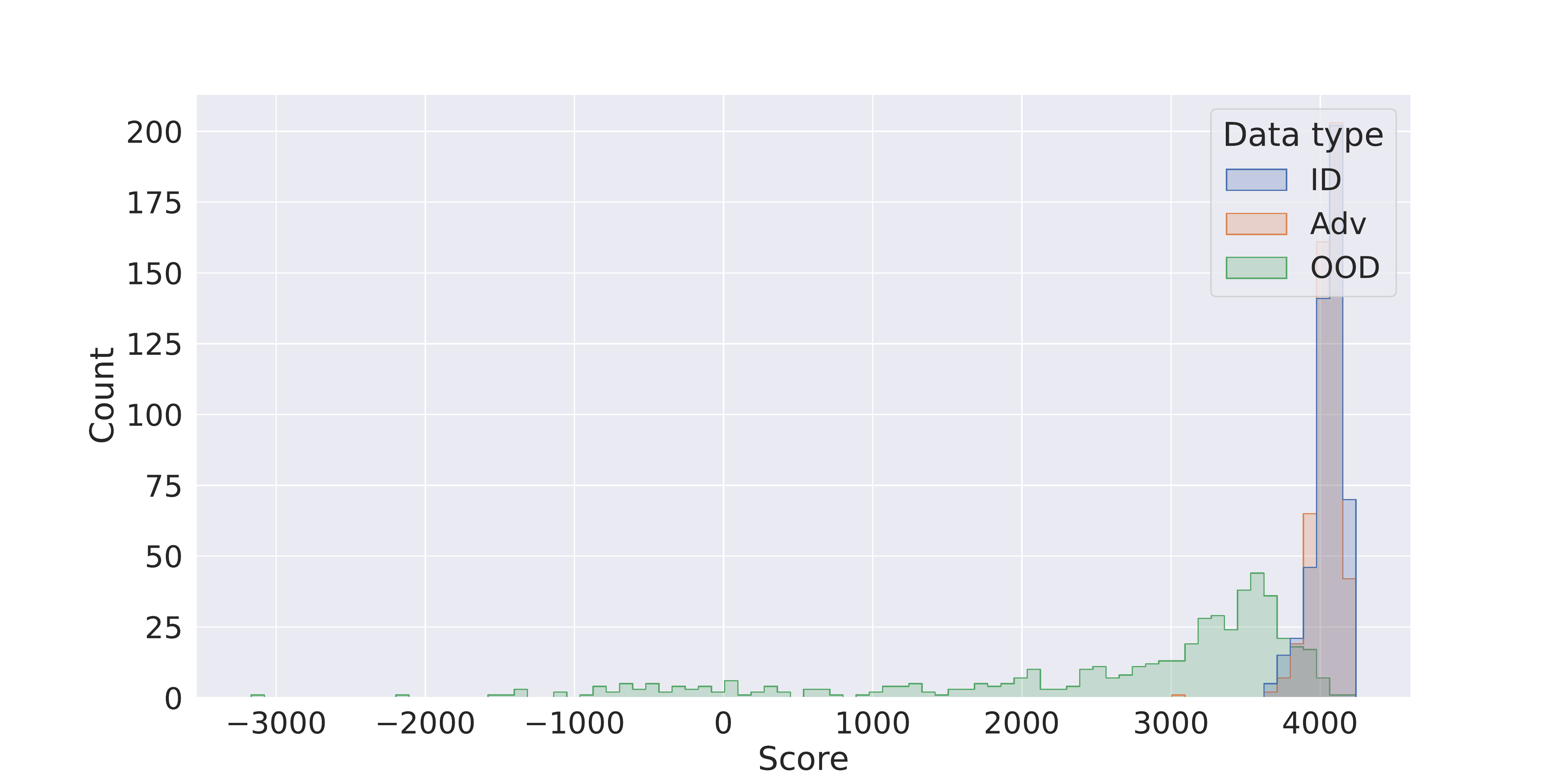}
\caption{Layer 2}
\label{fig:cls-layer-2-roberta-ag-pwws-imdb.pdf}
\end{center}
\end{subfigure}

\begin{subfigure}{0.42\textwidth}
\begin{center}
\includegraphics[trim=60 30 70 50, clip, width=0.95\textwidth]{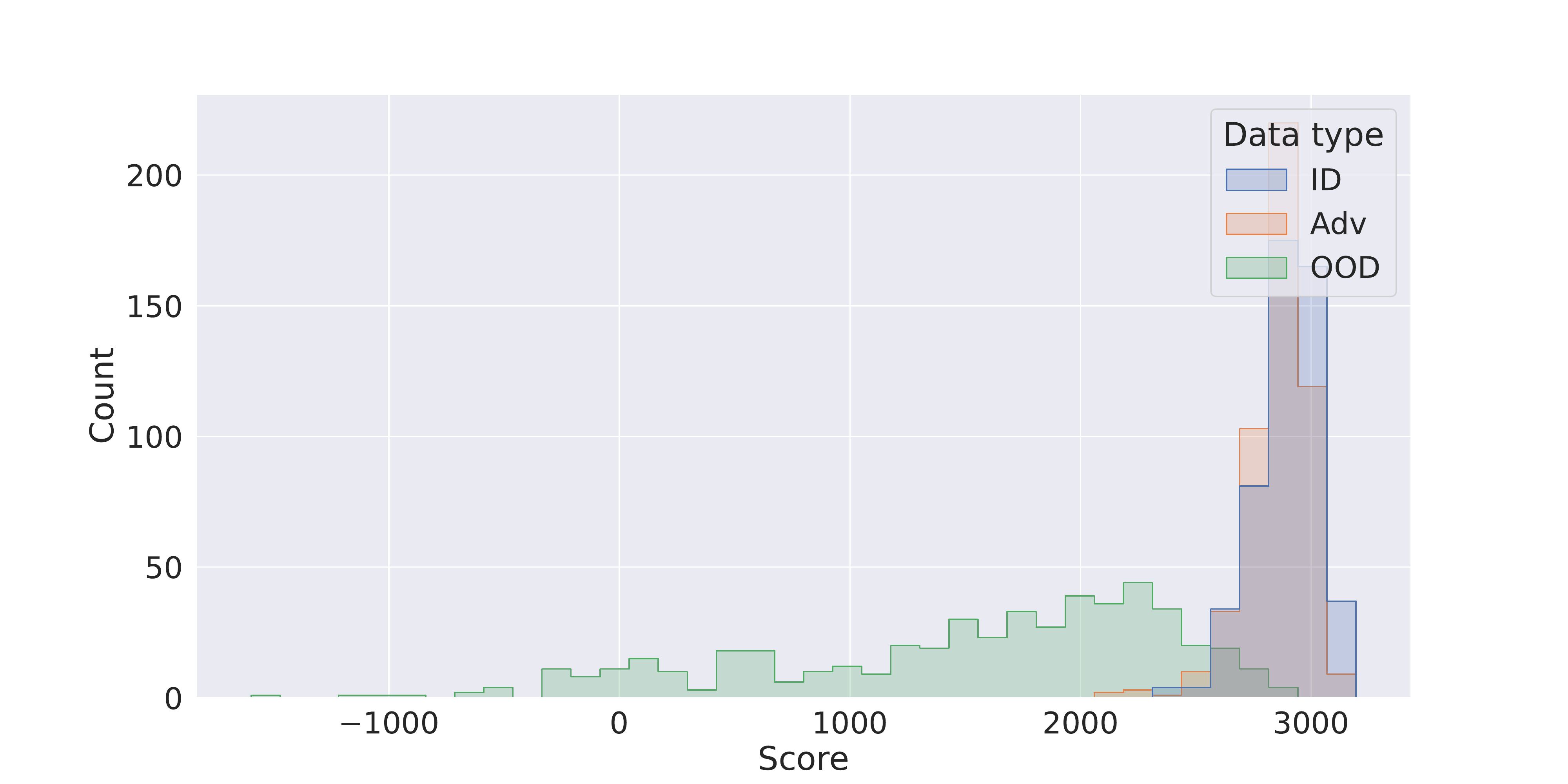}
\caption{Layer 4}
\label{fig:cls-layer-4-roberta-ag-pwws-imdb.pdf}
\end{center}
\end{subfigure}

\begin{subfigure}{0.42\textwidth}
\begin{center}
\includegraphics[trim=60 30 70 50, clip, width=0.95\textwidth]{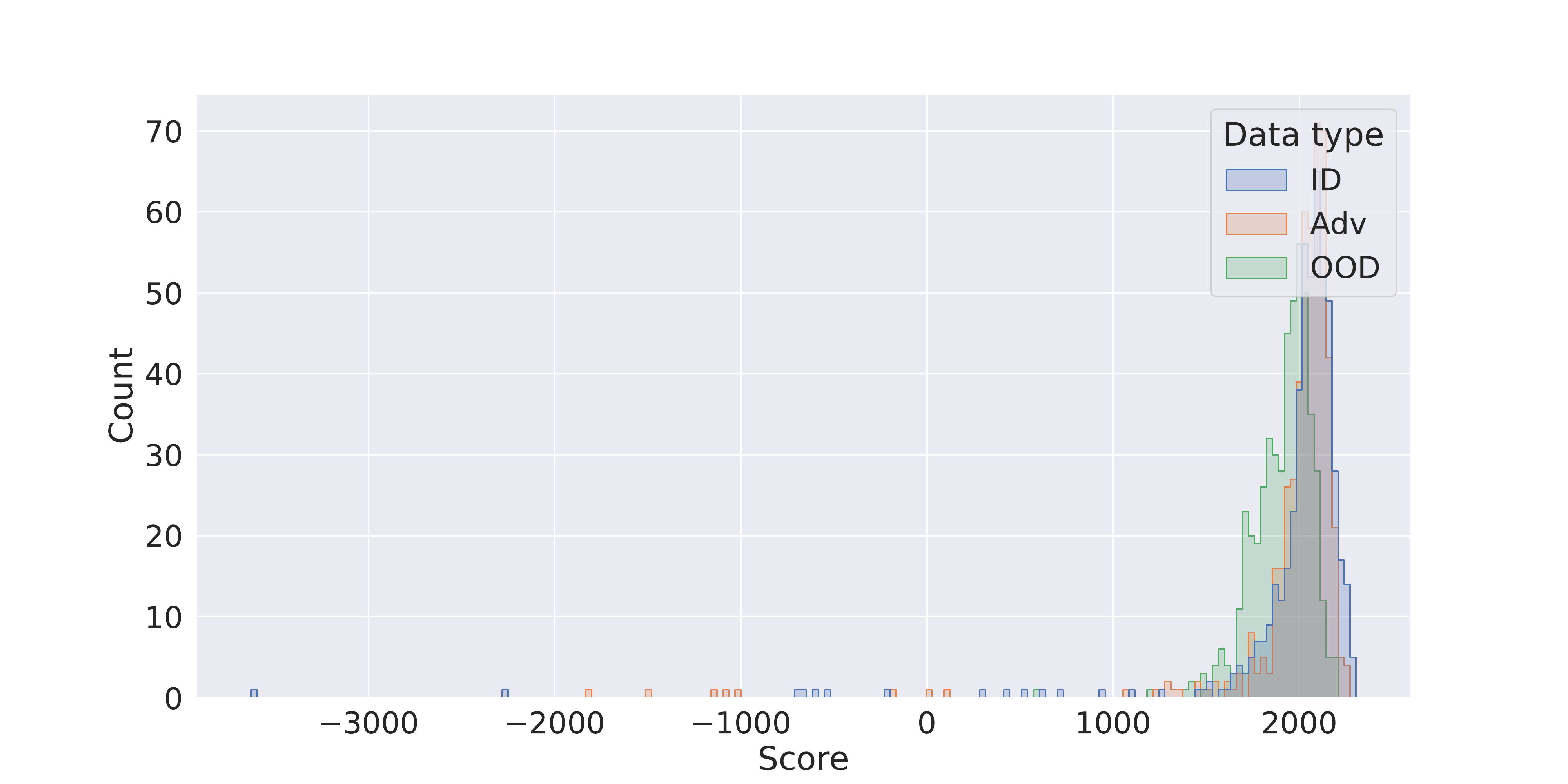}
\caption{Layer 6}
\label{fig:cls-layer-6-roberta-ag-pwws-imdb.pdf}
\end{center}
\end{subfigure}

\begin{subfigure}{0.42\textwidth}
\begin{center}
\includegraphics[trim=60 30 70 50, clip, width=0.95\textwidth]{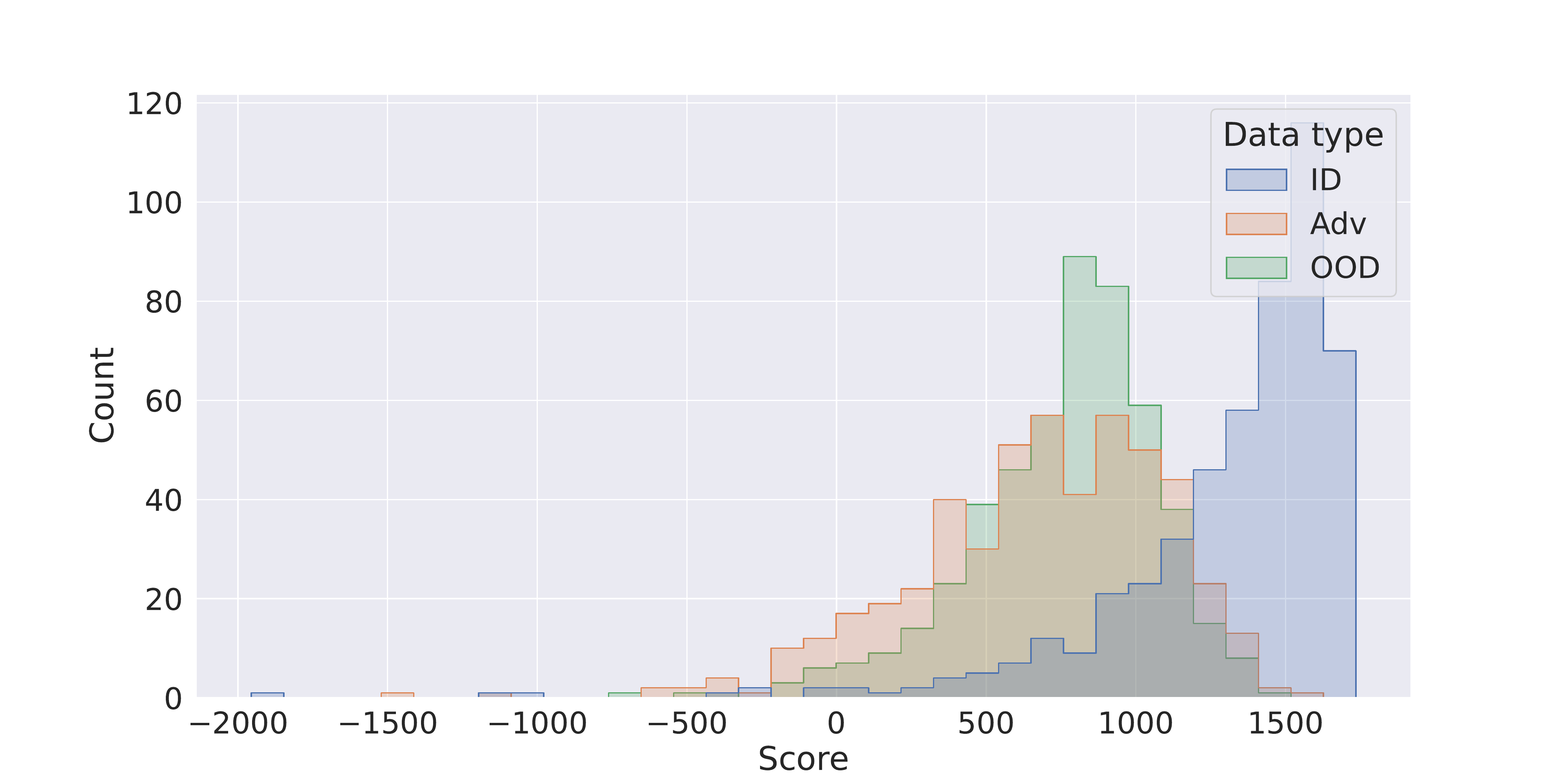}
\caption{Layer 8}
\label{fig:cls-layer-8-roberta-ag-pwws-imdb.pdf}
\end{center}
\end{subfigure}

\begin{subfigure}{0.42\textwidth}
\begin{center}
\includegraphics[trim=60 30 70 50, clip, width=0.95\textwidth]{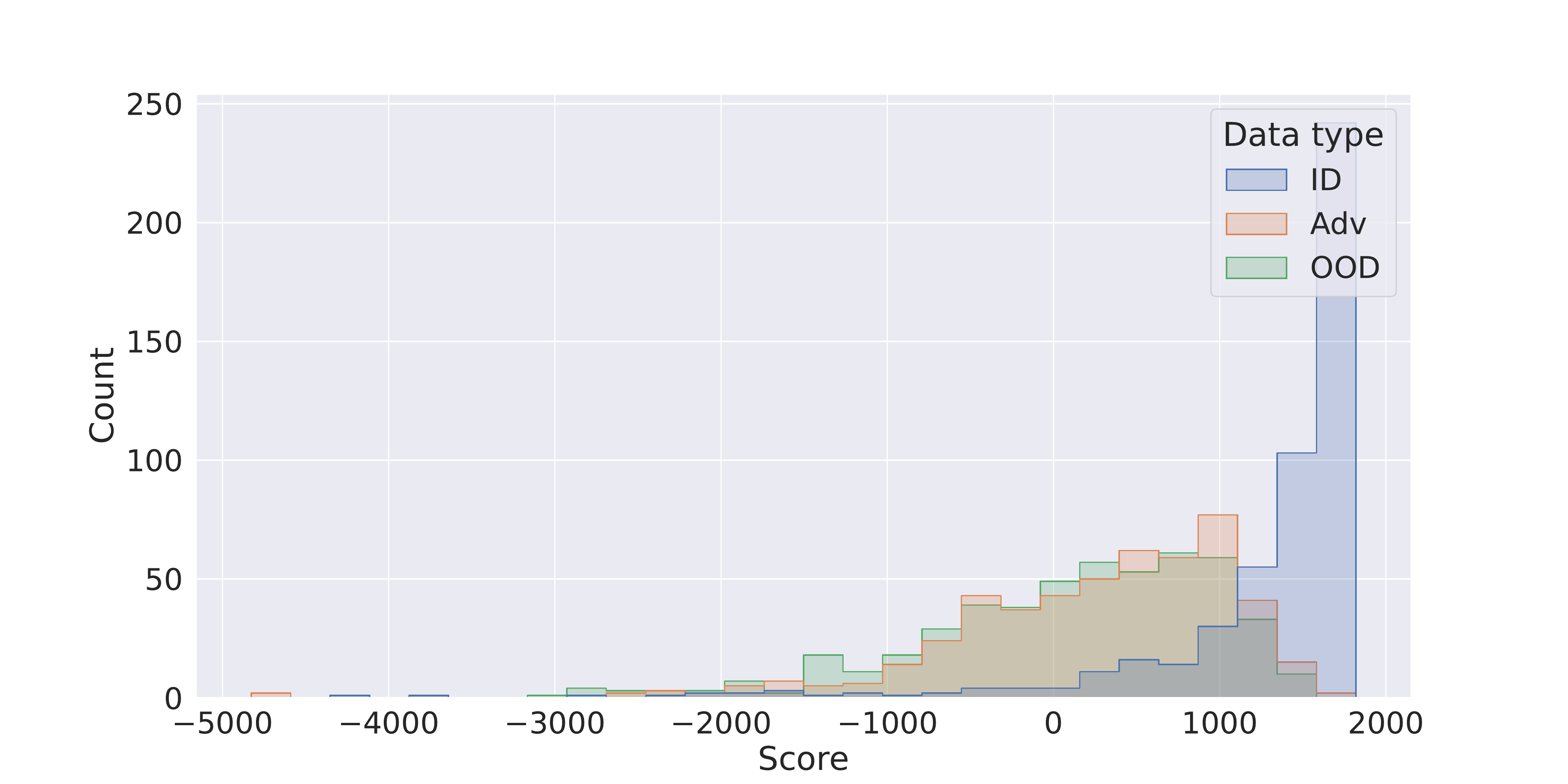}
\caption{Layer 10}
\label{fig:cls-layer-10-roberta-ag-pwws-imdb.pdf}
\end{center}
\end{subfigure}

\begin{subfigure}{0.42\textwidth}
\begin{center}
\includegraphics[trim=60 30 70 50, clip, width=0.95\textwidth]{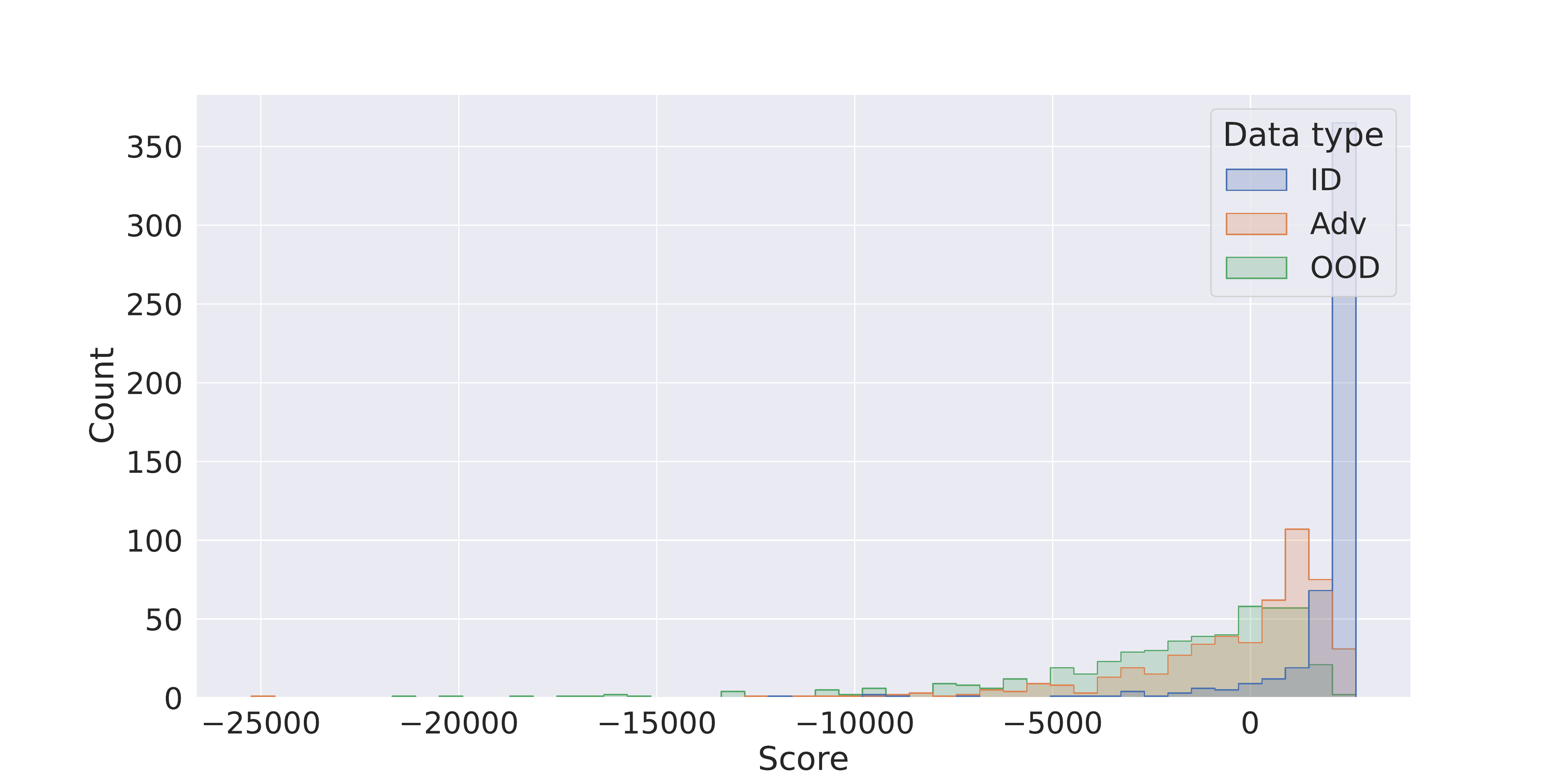}
\caption{Layer 12}
\label{fig:cls-layer-12-roberta-ag-pwws-imdb.pdf}
\end{center}
\end{subfigure}

\caption{The score $S^{l}(\mathbf{x}_{i})$ distribution, aggregation: $h_{i,CLS}^{l}$.
$\mathcal{D}_{ID,test}$: AG-News, $\mathcal{D}_{OOD}$: IMDB, $\mathcal{D}_{Adv}$: PWWS, model: RoBERTa. 
}
\label{fig:roberta mle distribution}
\end{figure}

In Figure~\ref{fig:roberta mle distribution}, we provide an illustration of how the score distributions based on the hidden feature in Section~\ref{subsec:hidden representations} evolve over the layers for a different combination of ID, OOD, Adv datasets.
In Figure~\ref{fig:roberta mle distribution}, the text classifier is fine-tuned from RoBERTa; this shows that our results are general across different pre-trained models.

\begin{figure}[t!]

\begin{subfigure}{0.45\textwidth}
\begin{center}
\includegraphics[trim=70 0 70 10, clip, width=0.95\textwidth]{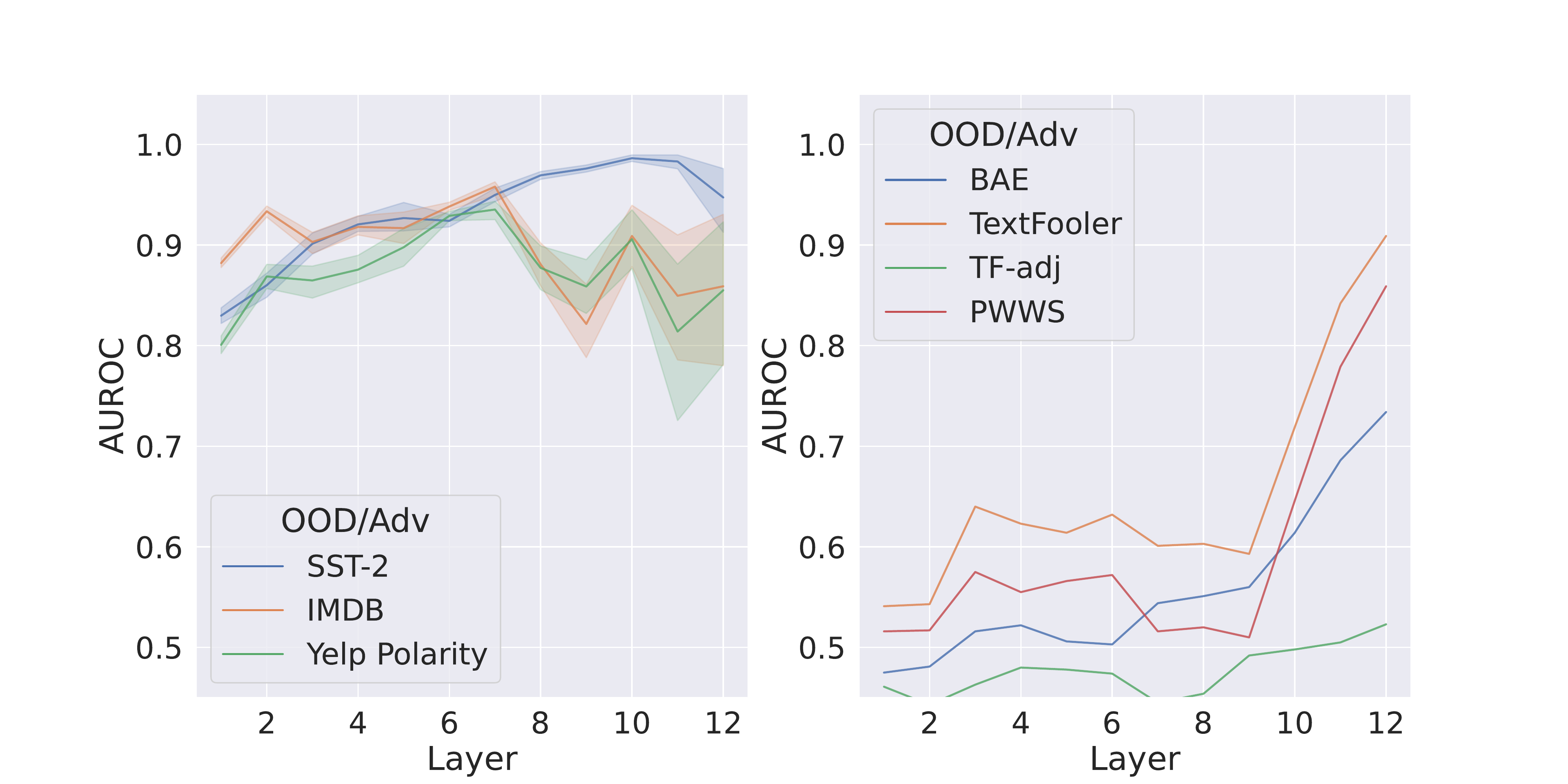}
\caption{$\mathcal{D}_{ID,test}$: AG-News, model: BERT}
\label{fig:cls-bert-ag-aucroc-2-types}
\end{center}
\end{subfigure}

\begin{subfigure}{0.45\textwidth}
\begin{center}
\includegraphics[trim=70 0 70 10, clip, width=0.95\textwidth]{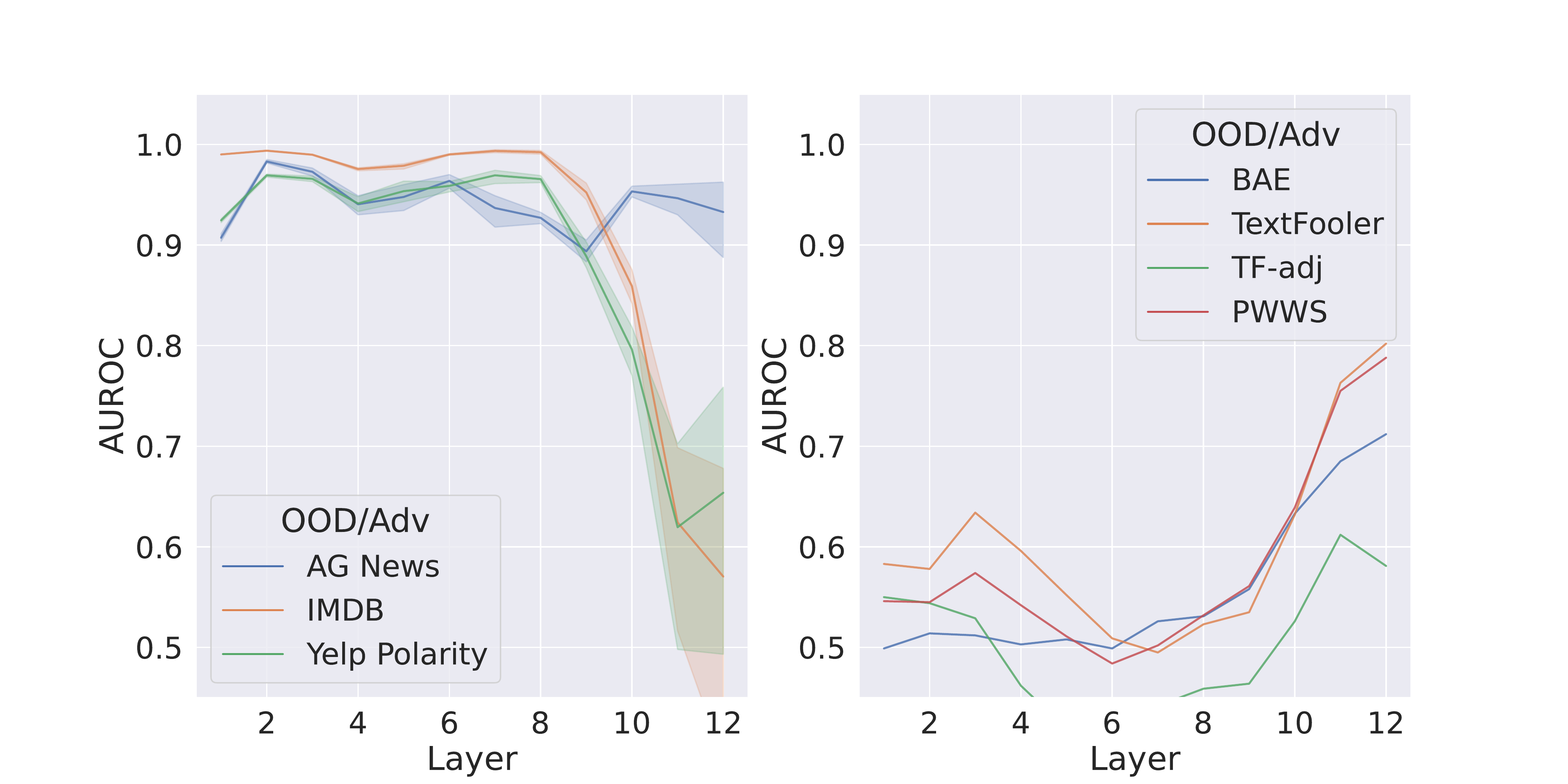}
\caption{$\mathcal{D}_{ID,test}$: SST-2, model: BERT}
\label{fig:cls-bert-sst2-aucroc-2-types}
\end{center}
\end{subfigure}

\begin{subfigure}{0.45\textwidth}
\begin{center}
\includegraphics[trim=70 0 70 10, clip, width=0.95\textwidth]{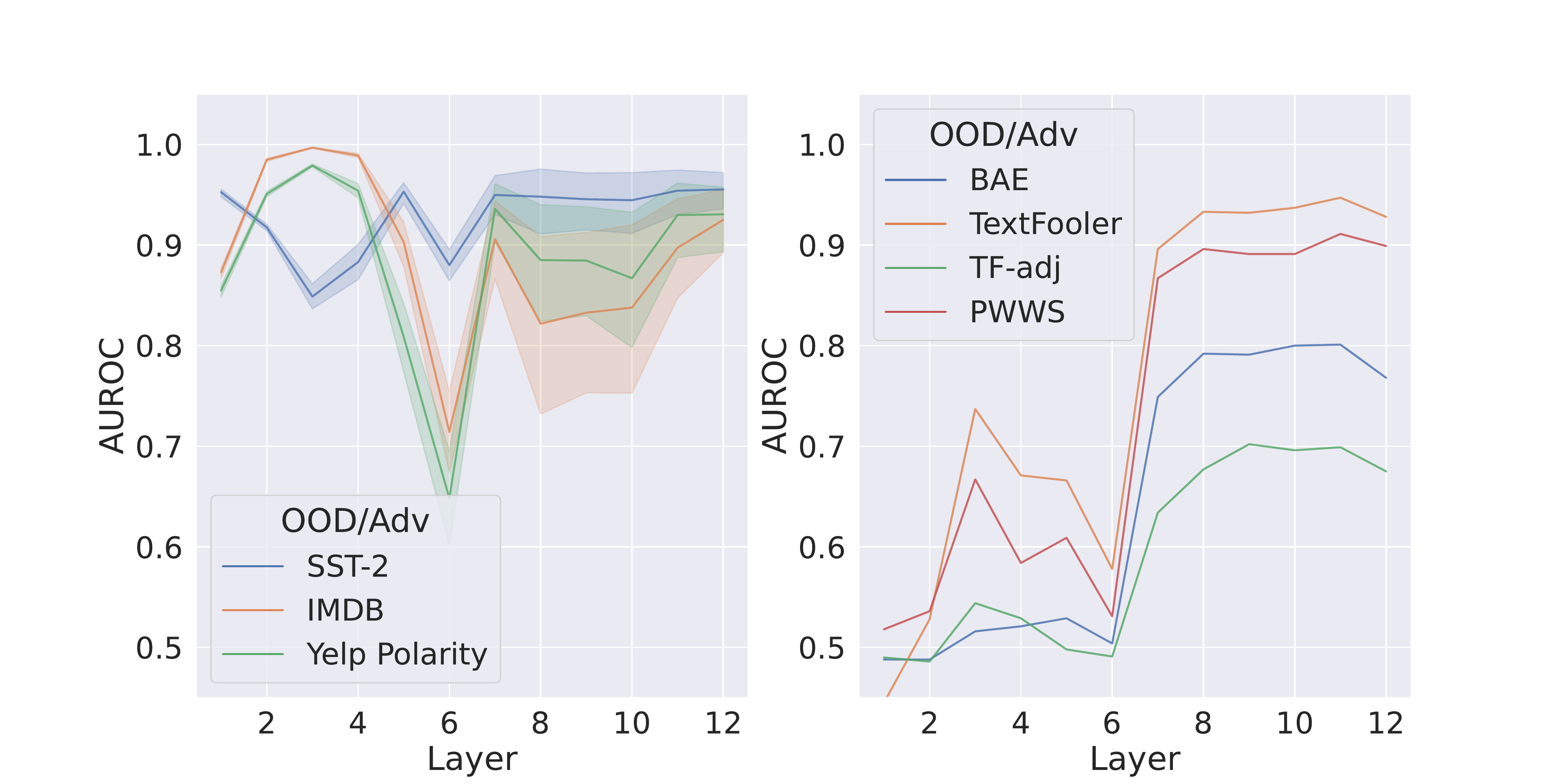}
\caption{$\mathcal{D}_{ID,test}$: AG-News, model: RoBERTa}
\label{fig:cls-roberta-ag-aucroc-2-types}
\end{center}
\end{subfigure}

\begin{subfigure}{0.45\textwidth}
\begin{center}
\includegraphics[trim=70 0 70 10, clip, width=0.95\textwidth]{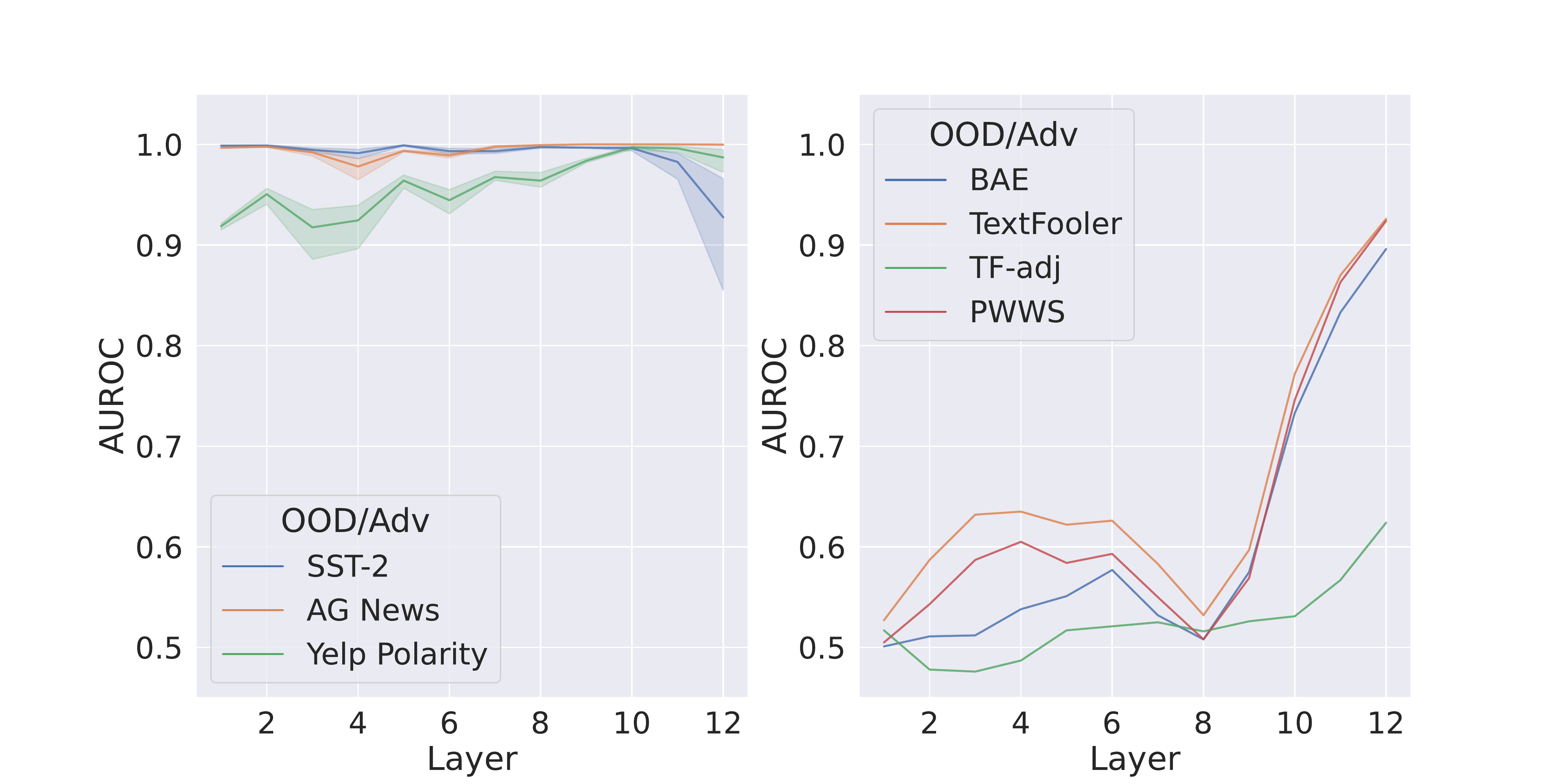}
\caption{$\mathcal{D}_{ID,test}$: IMDB, model: RoBERTa}
\label{fig:cls-roberta-imdb-aucroc-2-types}
\end{center}
\end{subfigure}

\begin{subfigure}{0.45\textwidth}
\begin{center}
\includegraphics[trim=70 0 70 10, clip, width=0.95\textwidth]{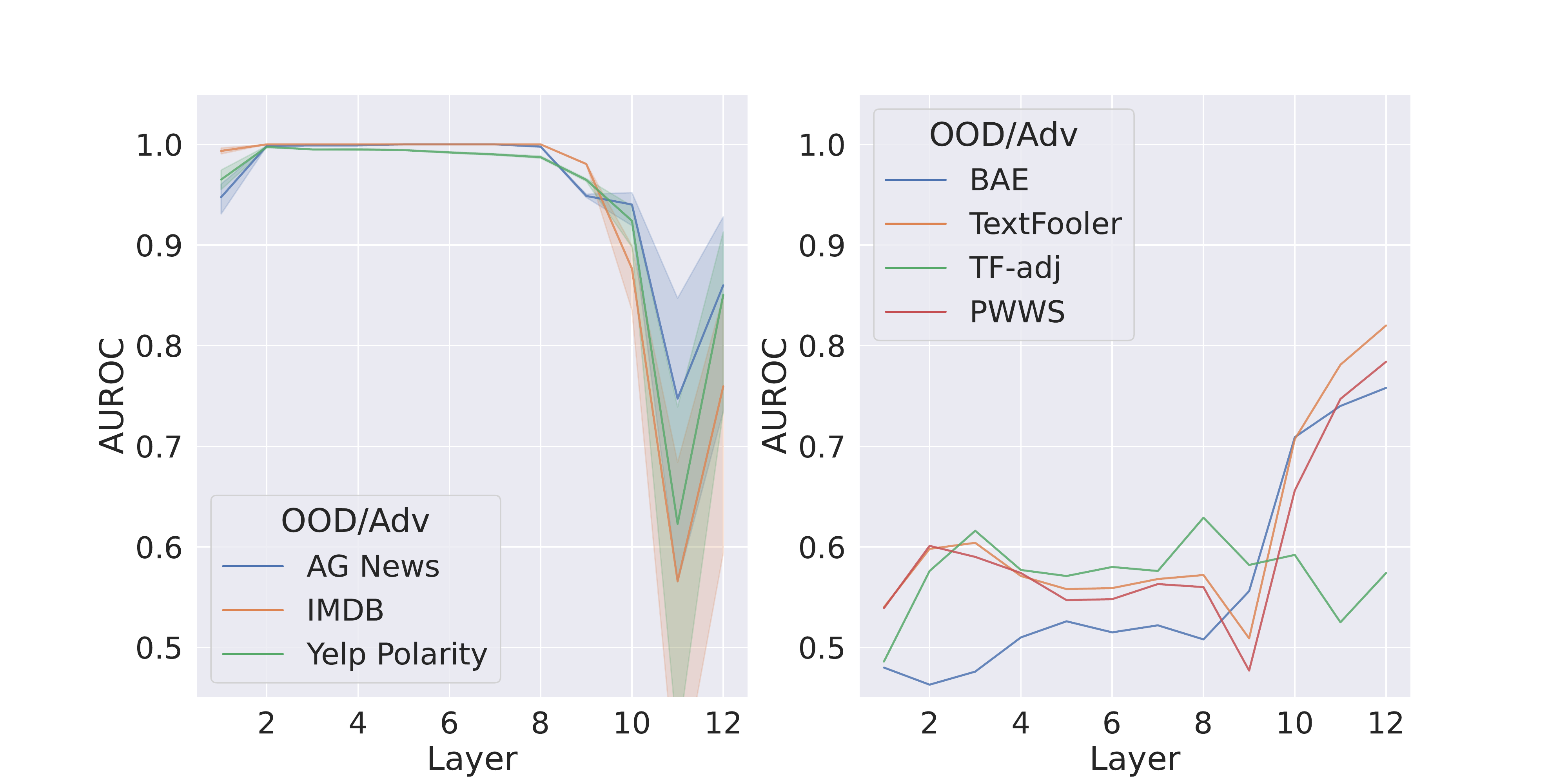}
\caption{$\mathcal{D}_{ID,test}$: SST-2, model: RoBERTa}
\label{fig:cls-roberta-sst2-aucroc-2-types}
\end{center}
\end{subfigure}

\caption{
Left: The detection results to separate $\mathcal{D}_{OOD}$ from $\mathcal{D}_{ID,test}$.
Right: The detection results to separate $\mathcal{D}_{Adv}$ from $\mathcal{D}_{ID,test}$.
Aggregation method: $h_{i,CLS}^{l}$.
}
\label{fig:app detect}
\end{figure}

Figure~\ref{fig:app detect} is the supplementary figure for Figure~\ref{fig:cls-bert-imdb-aucroc-2-types}.
From the left-hand side of all figures, we can find that all OOD datasets illustrate similar behaviors for all models: It is generally harder to differentiate between OOD and ID in the earlier layers of the model.
When the layers get deeper, some OOD datasets become inseparable with ID datasets while some can still be easily separated from by the features.
The behavior of Adv datasets, on the right-hand side, is also similar to that we presented in Figure~\ref{fig:cls-bert-imdb-aucroc-2-types}: Adv datasets are similar with ID samples in the shallower layers, and they grow more different with ID samples abruptly in the deeper layers.
While the detection results for TF-adj is poor for some datasets, we remind the readers that this is mainly due to the size of $\mathcal{D}_{Adv}$ are much smaller, as noted in Footnote~\ref{footnote: tf-adj}.
Refer to Table~\ref{tab:N} for the $N$ for different $\mathcal{D}_{Adv}$.

\begin{table}[ht]
    \caption{$N$ for different $\mathcal{D}_{Adv}$.
    }
    \label{tab:N}
    \centering
    \begin{tabular}{c|cccc}
        \hline
        \diagbox[width=6em]{Adv}{ID} & IMDB & SST-2 & AG-News\\
        \hline \hline
        \multicolumn{4}{c}{BERT} \\
        \hline
        TextFooler & 500 & 500 & 500\\
        PWWS & 500 & 500 & 500 \\
        BAE & 500 & 500 & 500 \\
        TF-adj & 500 & \textbf{40} & \textbf{172} \\
        \hline 
        \multicolumn{4}{c}{RoBERTa} \\
        \hline
        TextFooler & 500 & 500 & 500\\
        PWWS & 500 & 500 & 500 \\
        BAE & 500 & 500 & 500 \\
        TF-adj & \textbf{409} & \textbf{30} & \textbf{183} \\
        \hline
    \end{tabular}
    
\end{table}

\section{Supplementary Results for Section~\ref{sec: separate}}
\subsection{Stage 1 and Stage 2 Results of RoBERTa}
In this section, we show the stage 1 and stage 2 separation results for RoBERTa. 
The stage 1 detection result is in Table~\ref{tab:stage 1 roberta}, and the stage 2 detection result is in Table~\ref{tab:stage 2 roberta}. 
\begin{table}[h]
    \caption{The AUROC of detector for separating OOD samples from ID and Adv samples.
    The AUROC score for a entry is the averaged AUROC for a specific combination of \{$\mathcal{D}_{ID,test}$, $\mathcal{D}_{OOD}$\} when we vary the $\mathcal{D}_{Adv}$.
    The variance across different $\mathcal{D}_{Adv}$ is very small and thus not shown.
    The text classification model is fine-tuned from RoBERTa.
    }
    \label{tab:stage 1 roberta}
    \centering
    \begin{tabular}{c|ccc}
        \hline
        \diagbox[width=6em]{OOD}{ID} & IMDB & SST-2 & AG-News\\
        \hline \hline
        IMDB & - & 1.00& 0.99\\
        SST-2 & 1.00& -& 0.92 \\
        AG-News & 1.00 & 1.00 & - \\
        Yelp Polarity &0.95 & 1.00 & 0.95 \\
        \hline 
    \end{tabular}
    
\end{table}

\begin{table}[ht]
    \caption{The AUROCs of the detectors for separating Adv samples from ID ones.
    The text classification model is fine-tuned from RoBERTa.
    }
    \label{tab:stage 2 roberta}
    \centering
    \begin{tabular}{c|cccc}
        \hline
        \diagbox[width=6em]{Adv}{ID} & IMDB & SST-2 & AG-News\\
        \hline \hline
        TextFooler & 0.99 & 0.91 & 0.93\\
        PWWS & 0.99 &0.90 & 0.95 \\
        BAE & 0.98 & 0.78 & 0.92 \\
        TF-adj &0.87 & 0.79 & 0.84 \\
        \hline 
    \end{tabular}
    
\end{table}

\subsection{Experiments for Cascading Stage 1 and Stage 2}
\label{app: Experiments for Cascading Stage 1 and Stage 2}
In this section, we explain how we cascade stage 1 and stage 2.
\subsubsection{Experiment Setup}
\label{app: 1 + 2 setup}
We have a trained text classifier and the corresponding $\mathcal{D}_{ID, train}$, and we want to separate OOD, Adv, and ID samples using some threshold-based detectors, instead of training a classifier to distinguish different types of samples.
We also assume we have a in-distribution development set, $\mathcal{D}_{ID, dev}$, for selecting the thresholds of the classifier.
We select the threshold of stage 1 using the following procedure:
First, we calculate the score $S^2(\mathbf{x}_{i})$, the score calculated based on the hidden representations from the second layer, of each samples in $\mathcal{D}_{ID, dev}$.
The from the set $\mathbf{S}_1 = \{S^2(\mathbf{x}_{i})|\mathbf{x}_i\in \mathcal{D}_{ID, dev}\}$, we set the threshold $t_1$ by the 5-th percentile of $\mathbf{S}_1$.
That is, there will be 5\% of instances in $\mathcal{D}_{ID, dev}$ whose $S^2(\mathbf{x}_{i})$ is less than or equal to $t_1$.

In stage 2, the threshold $t_2$ is set in a similar way.
We collect the maximum output probability of every instances in $\mathcal{D}_{ID, dev}$ to form the set $\mathbf{S}_2$ and set $t_2$ to the 5-th percentile of $\mathbf{S}_2$.
Note that in this setting, we do not require any knowledge of the Adv and OOD datasets.
Also note that in our setting, for a single ID dataset, the $t_1$ for all OOD datasets is the same, and the $t_2$ for all kinds of adversarial attacks is the same.
The performance can sure be improved if one have the knowledge of OOD and Adv datasets, and the 5-th percentile may not be the best way for selecting the threshold.
Selecting the 5-th percentile as the threshold means that we think the 5\% that are the least similar with ID's development samples should be considered abnormal.

When testing the detectors, we sample 500 instances from  $\mathcal{D}_{ID,test}$, $\mathcal{D}_{OOD}$, and $\mathcal{D}_{Adv}$, except those datasets that are too small to sample 500 instances.
Since separating these three types of datasets is just a 3-way classification problem, we simply report the accuracy.
\subsubsection{Results}
We show the results in Table~\ref{tab:bert imdb 1 + 2},~\ref{tab:bert sst2 1 + 2}, and~\ref{tab:bert ag 1 + 2}, those results are based on text classifiers fine-tuned from BERT.
Is is obvious that the results of our proposed method perform far better than random guessing, which should give 33\% accuracy in case when all three types of datasets have 500 instances.
\begin{table}[ht]
    \caption{The accuracy for separating ID, Adv, and OOD samples, ID is IMDB.
    Yelp is short for Yelp Polarity.
    }
    \label{tab:bert imdb 1 + 2}
    \centering
    \begin{tabular}{c|ccc}
        \hline
        \diagbox[width=6em]{Adv}{OOD} & AG-News & SST-2 & Yelp\\
        \hline \hline
BAE     &    0.890   &  0.890   &      0.784     \\
PWWS    &     0.909  &   0.909   &      0.808    \\
TextFooler &  0.914  &   0.914    &     0.831    \\
TF-adj  & 0.722 &    0.722      &   0.634\\
        \hline 
    \end{tabular}
    
\end{table}
\begin{table}[ht]
     \caption{The accuracy for separating ID, Adv, and OOD samples, ID is SST-2.
    Yelp is short for Yelp Polarity.
    }
    \label{tab:bert sst2 1 + 2}
    \centering
    \begin{tabular}{c|ccc}
        \hline
        \diagbox[width=6em]{Adv}{OOD} & AG-News & IMDB & Yelp\\
        \hline \hline
BAE     &    0.644 & 0.671     &    0.618      \\
PWWS    &    0.739  & 0.760    &     0.714    \\
TextFooler & 0.784 & 0.809     &    0.762    \\
TF-adj  & 0.889 & 0.918    &     0.850\\
        \hline 
    \end{tabular}
   
\end{table}
\begin{table}[ht]
    \caption{The accuracy for separating ID, Adv, and OOD samples, ID is AG-News.
    Yelp is short for Yelp Polarity.
    }
    \label{tab:bert ag 1 + 2}
    \centering
    \begin{tabular}{c|ccc}
        \hline
        \diagbox[width=6em]{Adv}{OOD} & IMDB & SST-2 & Yelp\\
        \hline \hline
BAE     &    0.542 &  0.634   &      0.533     \\
PWWS    &     0.669 & 0.744   &      0.661   \\
TextFooler & 0.747 & 0.823   &      0.751    \\
TF-adj  & 0.565 & 0.704     &    0.553 \\
        \hline 
    \end{tabular}
    
\end{table}

\subsubsection{Setting the Thresholds in Stage 1 and Stage 2}
\label{app: Setting the Thresholds in Stage 1 and Stage 2}
\paragraph{With the Knowledge of OOD and Adv Datasets}
In this case, the thresholds can be easily selected.
Just find the optimal threshold for a pre-defined precision/recall score by considering the scores the ID, OOD, and Adv samples.
\paragraph{Without the Knowledge of OOD and Adv Datasets}
In this case, the thresholds can be determined by following the procedures in Appendix~\ref{app: 1 + 2 setup}.
\section{FAQ}
\label{app: faq}
\begin{itemize}
    \item [Q1] Why do you use the vocabulary of bert-base-uncased and roberta-base for the BOW features in Section~\ref{sec: input features}?
    \item [A1] We choose these two vocabulary sets for better comparison with the experiments that utilize features extracted from the text classifiers fine-tuned from BERT and RoBERTa, whose input are the sentences tokenized by the two tokenizers. 
    \item [Q2] Have you tried other methods for comparing the BOW feature's similarity with the ID ones?
    \item [A2] Given a sentence and its BOW feature, we also try to calculate the cosine similarity with each BOW feature in $\mathcal{D}_{ID, train}$ and take the maximum one.
    We find the results using this method will mostly be inferior to the method we used in Section~\ref{sec: input features}.
    \item [Q3] How to construct a single OOD detector for all kinds of OOD datasets for stage 1 in Section~\ref{sec: separate}?
    \item [A3] Consider we want to separate OOD samples from ID samples and Adv samples, where the ID samples are from IMDB.
    Since all types of Adv look exactly like ID samples in earlier layers, we do not need to build detectors for all combinations of $\mathcal{D}_{OOD}$, $\mathcal{D}_{Adv}$ for a fixed $\mathcal{D}_{ID,test}$; we just need to build detectors for different $\mathcal{D}_{OOD}$ for a fixed $\mathcal{D}_{ID,test}$.
    We build the OOD detector to detect SST-2, the OOD detector for AG-News, and the OOD detector for Yelp Polarity.
    These three detectors differ in how they set the threshold of labeling the inputs as positive based on their score.
    What we need to do is to select the highest threshold among the previous three thresholds, and based on Table~\ref{tab:stage 1}, we might want to use the threshold of the detector for Yelp Polarity.
    \item [Q4] Where does the variance in the left-hand side of Figure~\ref{fig:cls-bert-imdb-aucroc-2-types} come from?
    \item [A4] It is the variance of sampling different $\mathcal{D}_{ID,test}$.
    In our implementations, for the same ID dataset, the $\mathcal{D}_{ID,test}$ for different $\mathcal{D}_{Adv}$ may consist different instances.
    Considering a benign testing sample in the testing/development set of ID dataset, different attacking methods can successfully attack that instance while the attack may also fail.
    If the instance is successfully attacked, then its adversarial counterpart will be included in $\mathcal{D}_{Adv}$, and the original instance cannot be included in $\mathcal{D}_{ID,test}$ since we require $\mathcal{D}_{ID,test}$ and $\mathcal{D}_{Adv}$ to consist no benign/adversarial pairs.
    On the other hand, if an instance was not successfully attacked, then it can be included in the $\mathcal{D}_{ID,test}$.
    This makes different $\mathcal{D}_{Adv}$ to have slightly different $\mathcal{D}_{ID,test}$.
    \item [Q5] In Section~\ref{sec: separate}, why do the authors choose to use the maximum probability in Section~\ref{subsec: output probability distribution} as an indicator of Adv samples, instead of using the  score $S^{l}(\mathbf{x}_{i})$ based on the hidden representations in the deeper layers?
    Based on the results in Figure~\ref{fig:cls-bert-imdb-aucroc-2-types} and Figure~\ref{fig:app detect}, using $S^{12}(\mathbf{x}_{i})$ can also separate Adv and ID samples.
    \item [A5] While using $S^{12}(\mathbf{x}_{i})$ can also separate Adv samples from ID ones, we find that using the maximum probability from the output generally leads to better detection results.
    This can be observed from comparing Figure~\ref{fig:cls-bert-imdb-aucroc-2-types} and Figure~\ref{fig:app detect} with Table~\ref{tab:stage 2} and Table~\ref{tab:stage 2 roberta}.
    Also, when using $S^{12}(\mathbf{x}_{i})$, we also need to fit the features with a class-condition normal distribution, which requires additional parameters.
    Thus, we choose to use maximum probability due to its better performance and less parameters.
    \item [Q6] Why do the authors only include results of BERT in the paper?
    \item [A6] This is because the results for models fine-tuned from RoBERTa are like the results obtained from the text classifiers fine-tuned from BERT.
    \item [Q7] Isn't using the maximum probability to identify Adv samples unreasonable?
    Adv samples are made to make the model to be unconfident about, so it is odd to use it as an indicator as Adv samples.
    \item [A7] No, Adv samples are \textbf{NOT} designed to make the model unsure about their predictions. 
    Adv samples are crafted such the model makes the wrong prediction; the model can be very confident but wrong, it can also be unconfident and wrong.
    We show that the text classifier is unconfident and wrong on those Adv sample, and we use this as an indicator of Adv samples.
\end{itemize}

\end{document}